\documentclass[journal]{IEEEtran}

\usepackage{caption,subcaption,float}
\usepackage{graphicx}
\usepackage{multirow}
\usepackage{amsmath,fixmath,amsfonts}
\usepackage{array}
\usepackage{cite}
\usepackage{url}

\begin{document}
\title{A New Comprehensive Framework for Multi-Exposure Stereo Coding Utilizing Low Rank Tucker-ALS and 3D-HEVC Techniques}

\author{Mansi~Sharma$^{1}$,~Jyotsana Grover$^{2}$
\thanks{$^1$The author is with the Department
of Computer Science and Engineering, Thapar Institute of Engineering and Technology, Patiala, Punjab 147004;

E-mail: mansi.sharma@thapar.edu;

$^2$The author is with Department of Computer Science and Information Systems, Work Integrated Learning Programmes, Birla Institute of Technology and Science, Pilani, Pilani Campus

E-mail: \{jyotsana.grover@pilani.bits-pilani.ac.in\}
}}

\maketitle

\begin{abstract}
Display technology must offer high dynamic range (HDR) contrast-based depth induction and 3D personalization simultaneously. Efficient algorithms to compress HDR stereo data is critical. Direct capturing of HDR content is complicated due to the high expense and scarcity of HDR cameras. The HDR 3D images could be generated in low-cost by fusing low-dynamic-range (LDR) images acquired using a stereo camera with various exposure settings. In this paper, an efficient scheme for coding multi-exposure stereo images is proposed based on a tensor low-rank approximation scheme. The multi-exposure fusion can be realized to generate HDR stereo output at the decoder for increased realism and exaggerated binocular 3D depth cues.

For exploiting spatial redundancy/sparsity in LDR stereo images, the stack of multi-exposure stereo images is decomposed into a set of projection matrices and a core tensor following an alternating least squares Tucker decomposition model. The compact, low-rank representation of the scene, thus, generated is further processed by 3D extension of High Efficiency Video Coding standard. The encoding with 3D-HEVC enhance the proposed scheme efficiency by exploiting intra-frame, inter-view and the inter-component redundancies in low-rank approximated representation. We consider constant luminance property of IPT and $Y^{'}C_bC_r$ color space to precisely approximate intensity prediction and perceptually minimize the encoding distortion. Besides, the proposed scheme gives flexibility to adjust the bitrate of tensor latent components by changing the rank of core tensor and its quantization. Extensive experiments on natural scenes demonstrate that the proposed scheme outperforms state-of-the-art JPEG-XT and 3D-HEVC range coding standards.

\end{abstract}

\IEEEpeerreviewmaketitle

\section{Introduction}
\label{sec:intro}
The dynamic range of light intensity (luminance) is defined as the ratio of brightest luminance to the darkest luminance. The dynamic range of the human visual system in natural scene viewing can approach 20 stops, or 1,000,000:1, and often exceeds 10,000:1 \cite{IntroRef1}. However, contemporary displays are not capable of directly reproducing high dynamic range luminance-realistic images \cite{IntroRef2}. Solution based on improving display devices or digital camera hardware technology like the increasing sensitivity of lens, etc., can work to a certain extent only \cite{IntroRef3}. Besides, solutions designed to work for special conditions like static scene backgrounds, or sensing multiple images to generate HDR images, are not practically feasible \cite{IntroRef4, IntroRef5}. Computational photography, HDR imaging (HDRI) techniques are more feasible as they involve creating HDR image by modifying a captured signal LDR or SDR image or multiple images \cite{IntroRef6, IntroRef7, IntroRef8, IntroRef9, IntroRef10, IntroRef11, IntroRef12}. Computational HDRI techniques in general are categorized into methods involving capturing and combining multiple LDR images with varying exposures or single image based deep learning or manual processing approaches \cite{IntroRef6, IntroRef7, IntroRef8, IntroRef9, IntroRef10, IntroRef11, IntroRef12, IntroRef13, IntroRef14, IntroRef15, IntroRef16, IntroRef17}. The basic idea behind HDR image synthesis combining multiple images with varied illuminations is to identify relations between luminance intensities of LDR images and different exposures. The recovered non-linear camera response function that relates scene irradiance to image intensities is employed as a fundamental step in the generation of high dynamic image \cite{IntroRef9, IntroRef12}.

\begin{figure*}[t]
\centering
\includegraphics[height=50mm,width=180mm]{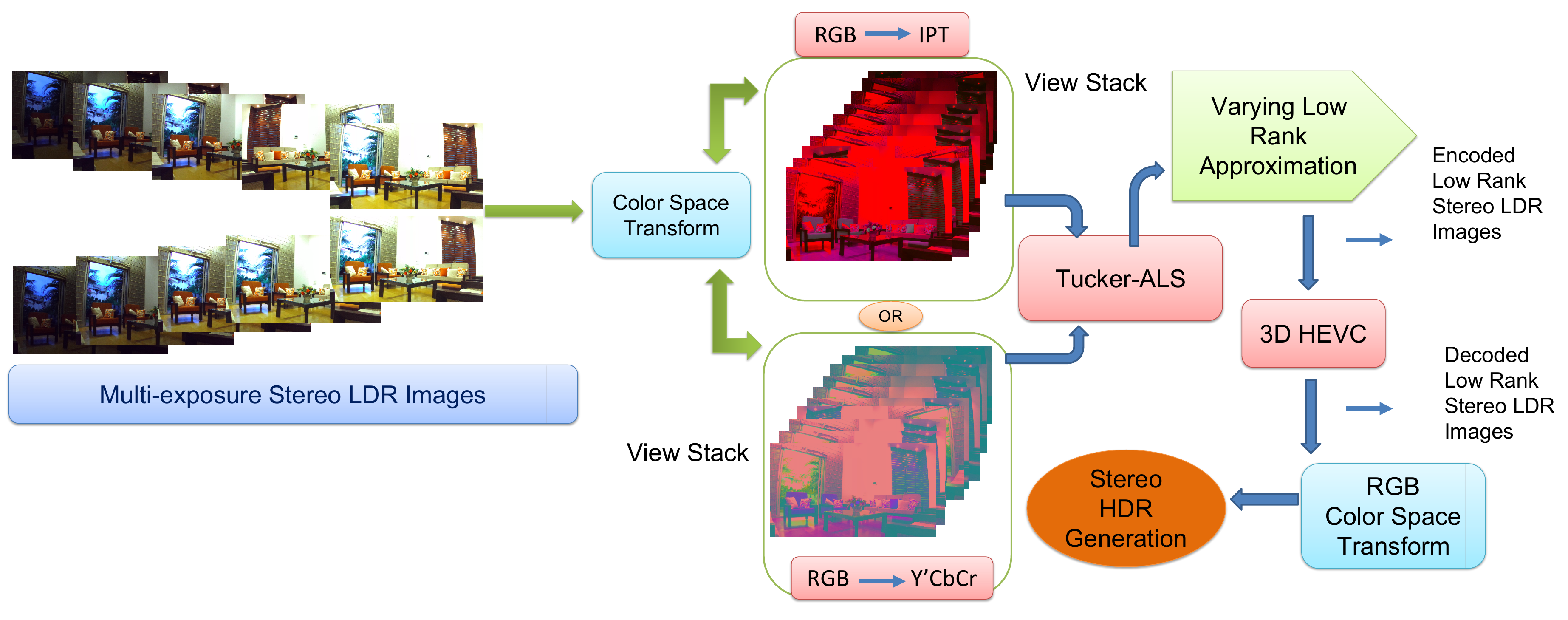}
\caption{Workflow of proposed coding scheme for multi-exposure stereo images and efficient HDR compression.}
\label{Fig1}
\end{figure*}

The fusion based HDR generation methods using single image are categorized into two groups: 1) manual methods based on filters, tone mapping, etc., 2) deep learning methods like CNNs, GANs, etc. In principle, such fusion methods are also based on multiple images with varied illuminations. The critical difference being that such
multiple LDR images are constructed internally by the algorithms itself. For example, the algorithms presented in \cite{IntroRef14, IntroRef15, IntroRef16} separate illumination and reflectance of the original image. The illumination image is basically a grayscale version of the original image, whereas the reflectance image describes its color characteristics. The main principle is to create images of various illuminations and combine them to generate an image which portray the higher illumination range of real world. Lastly, both illumination and reflectance images are combined to form the HDR image. Likewise, some approaches integrate scene details to generate an LDR image with enhanced contrast, properly weighted in the over-saturated and under-saturated regions. Nowadays, deep learning techniques produce convincing results recovering high luminance with colors and details of saturated areas using a single exposure image \cite{IntroRef6, IntroRef7, IntroRef13}. 

It is critical to efficiently compress and store HDR content with a wider intensity range. Radiance RGBE, Logluv TIFF, OpenEXR are popular file formats for HDR content storage in floating-point format. JPEG-XT, a new coding standard, addresses the requirements of directly coding images with higher bit depths (9 to 16 bits), HDR content, lossless compression, and representation of alpha channels \cite{IntroRef8}. JPEG-XT (ISO/IEC 18477) extends the JPEG specification in an absolutely backwards compatible manner \cite{Ref2}. H.264/MPEG-4 AVC and HEVC based coding algorithms convert HDR content into an integer format before encoding \cite{Ref2}. Besides, the non-backward compatible methods encode scene dynamic range employing perceptually motivated functions for chroma reproduction.

In this work, we propose a new paradigm for coding multi-exposure stereo images for 3D HDR display applications \cite{3DDISPLAY1, 3DDISPLAY2, 3DDISPLAY3}. More specifically, we proposed a Tucker Decomposition based encoding model with 3D-HEVC that efficiently exploits the intra-frame, inter-view, inter-component redundancies and simultaneously takes the sparsity/low-rankness of latent multi-expsoure stereo images representation into consideration. The key idea is to employ tucker decomposition, which decomposes the latent tensor representation of stereo LDR images into a set of projection matrices and a compact core tensor. Encoding low-rank approximated output with 3D-HEVC by varying the rank of core tensor and quantization levels, gives flexibility to adjust the bitrate of latent representation. Thus, a single integrated model is flexible to compress and reconstruct the decompressed images under multiple bitrates and quality levels.
Extensive experiments demonstrate that, our proposed compatible encoding scheme is competitive with state-of-the-art coding standards. It maintains good quality HDR imaging in terms of PSNR and HDR-VDP-2 visual quality with a range of bitrates.

The remainder of this paper is organized as follows: Section II
provides an overview of the state-of the-art related works. Section III presents the proposed multi-expsoure stereo coding scheme. Section IV shows the experimental results and analysis. Finally, in Section V, we draw conclusions and discuss the scope of future work.

\begin{figure*}[t] 
    \centering

\begin{tabular}{cccc}
\subfloat[ Garden(left)]{\includegraphics[width = 1.5in]{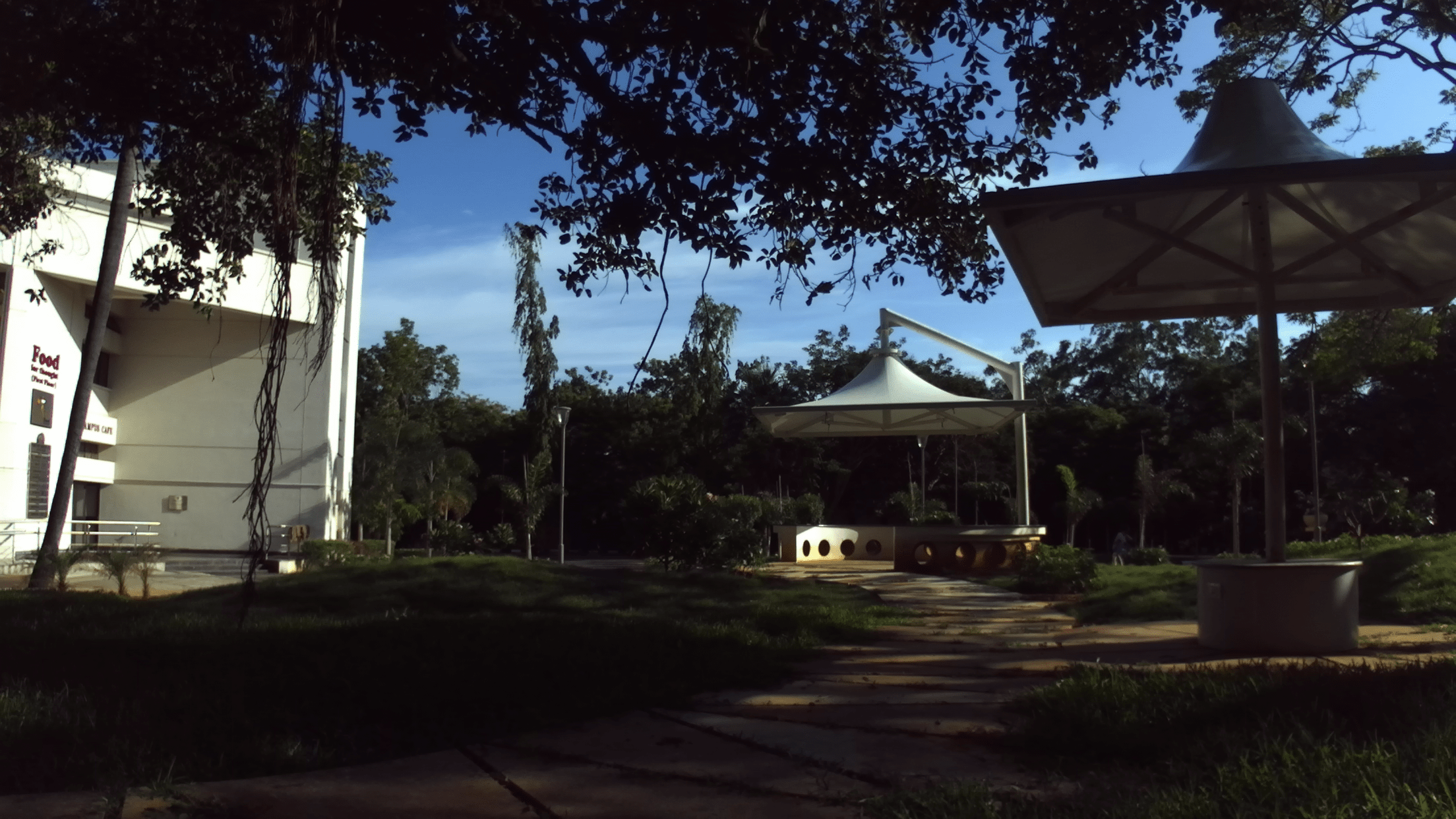}} &
\subfloat[Garden(right)]{\includegraphics[width = 1.5in]{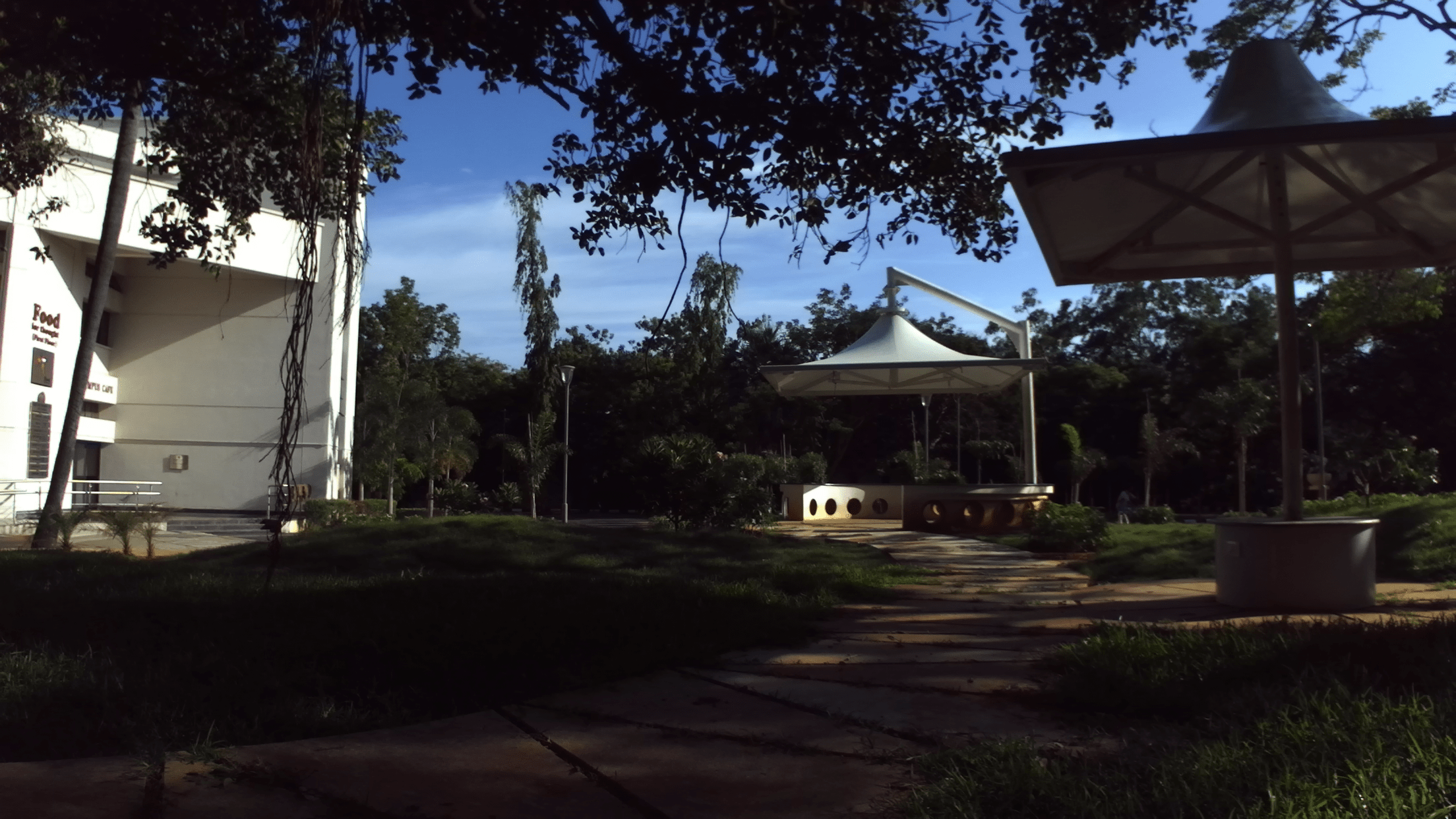}} &
\subfloat[ Garden(left)]{\includegraphics[width = 1.5in]{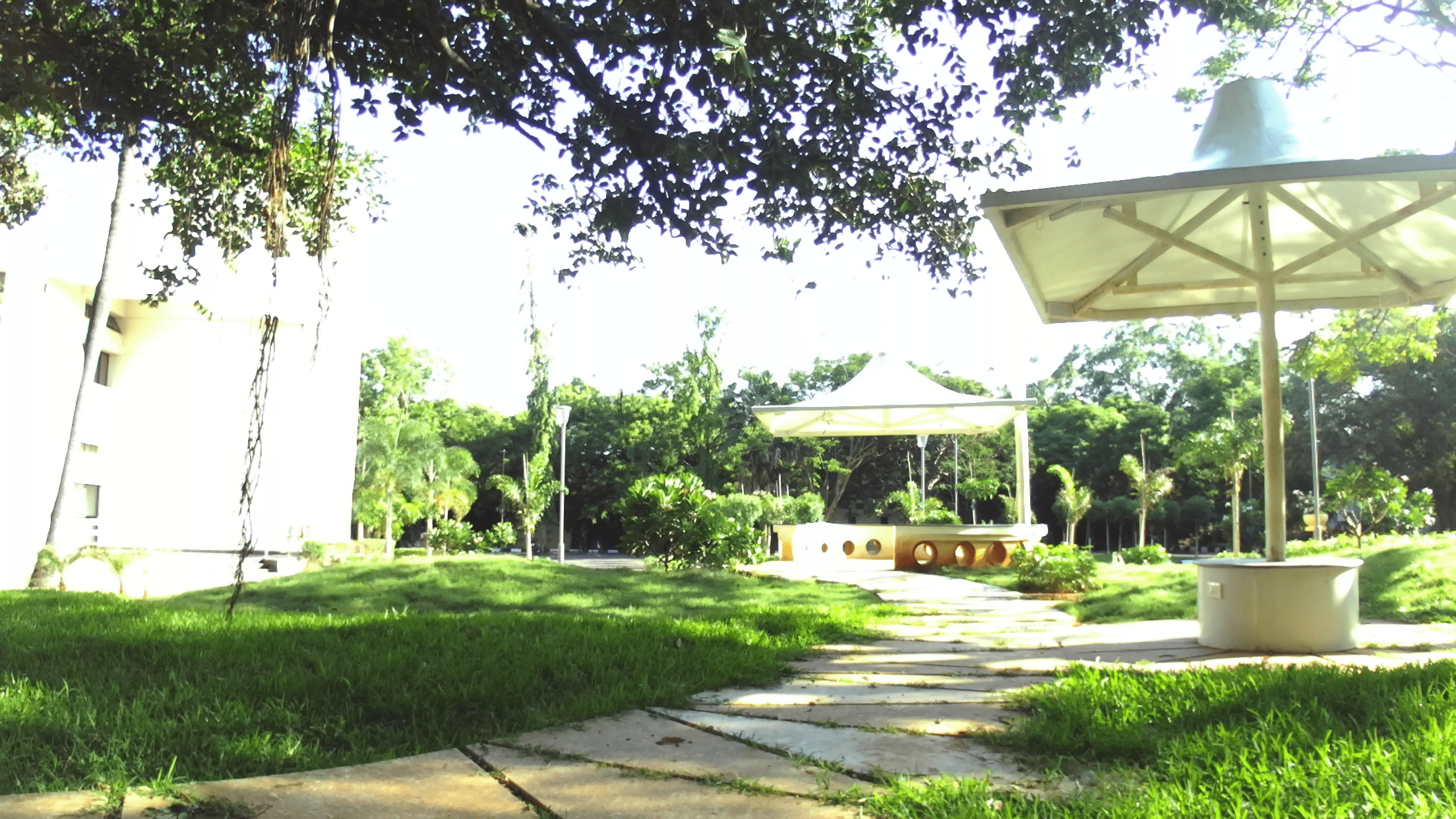}} &
\subfloat[ Garden(right)]{\includegraphics[width = 1.5in]{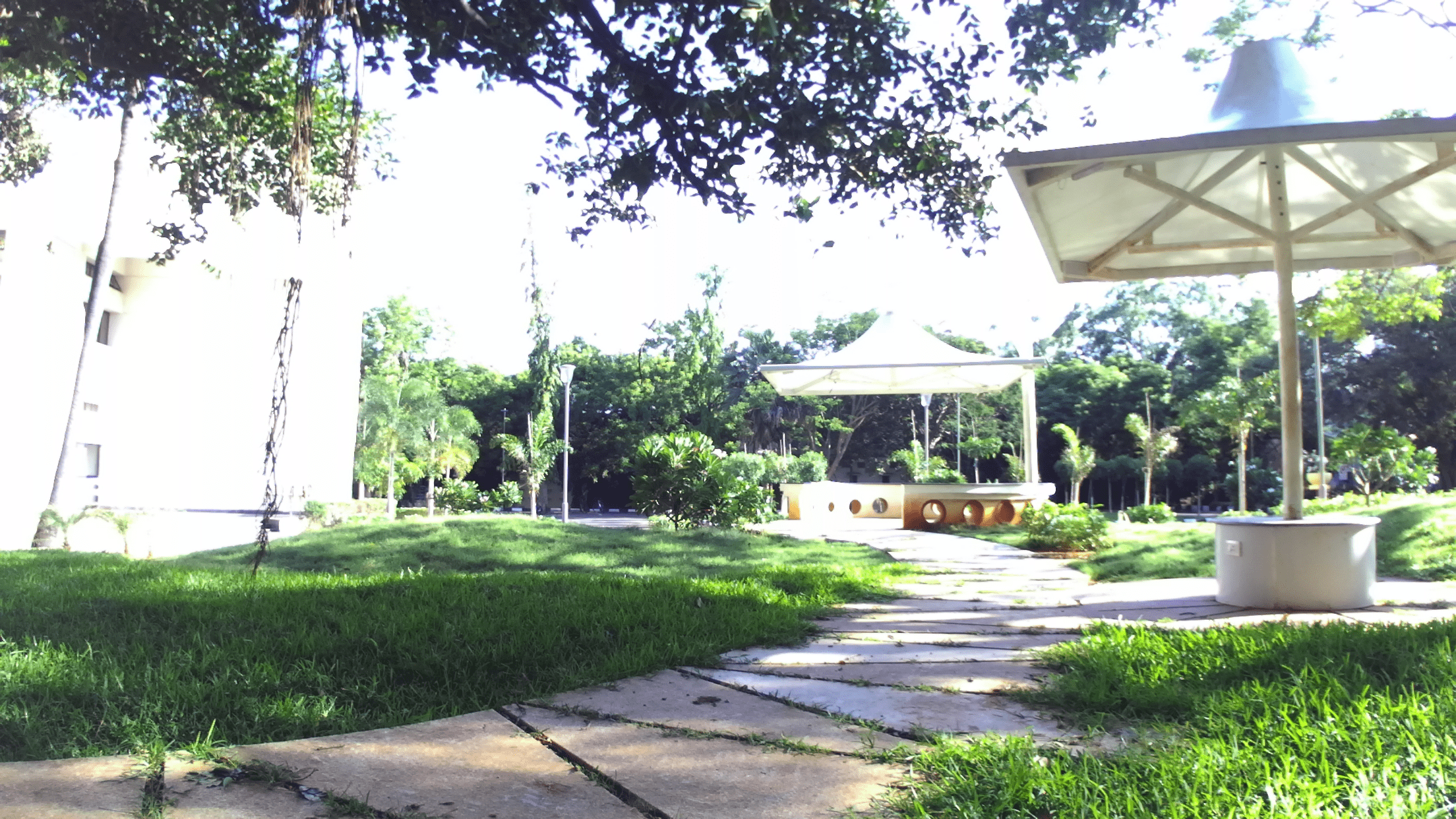}}
\\
\subfloat[ Gate(left)]{\includegraphics[width = 1.5in]{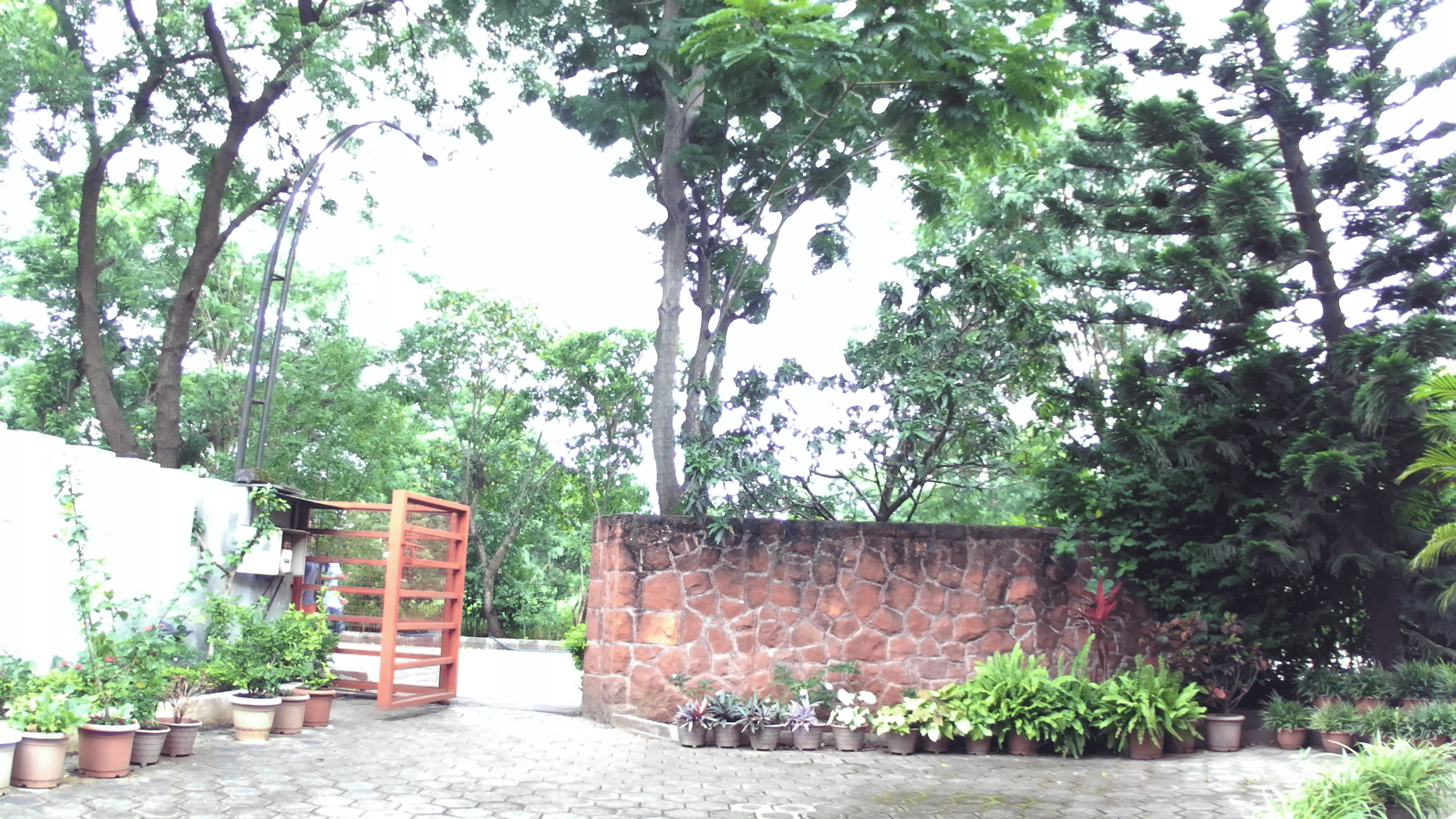}} &
\subfloat[ Gate(right)]{\includegraphics[width = 1.5in]{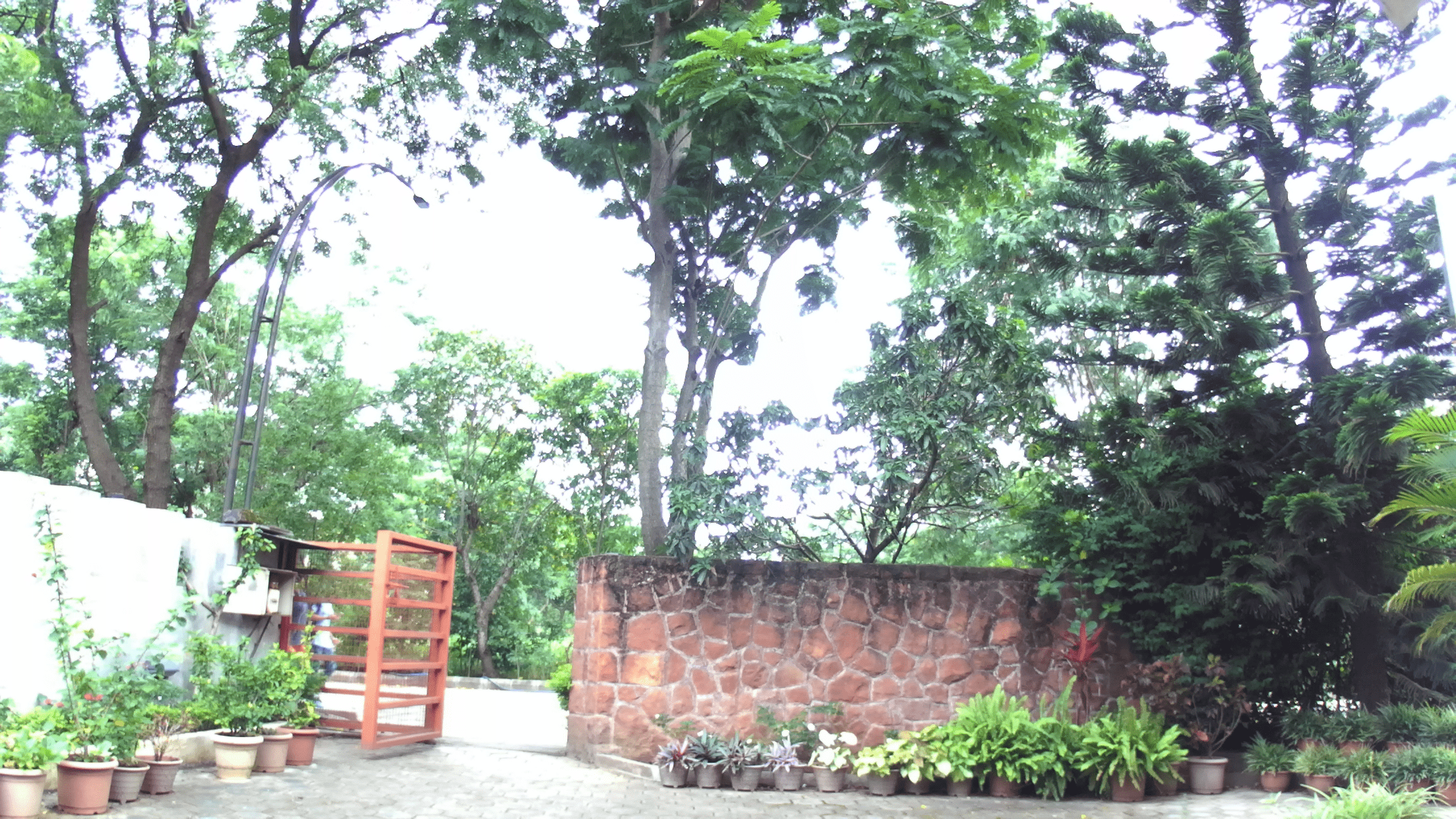}} &
\subfloat[ Gate(left)]{\includegraphics[width = 1.5in]{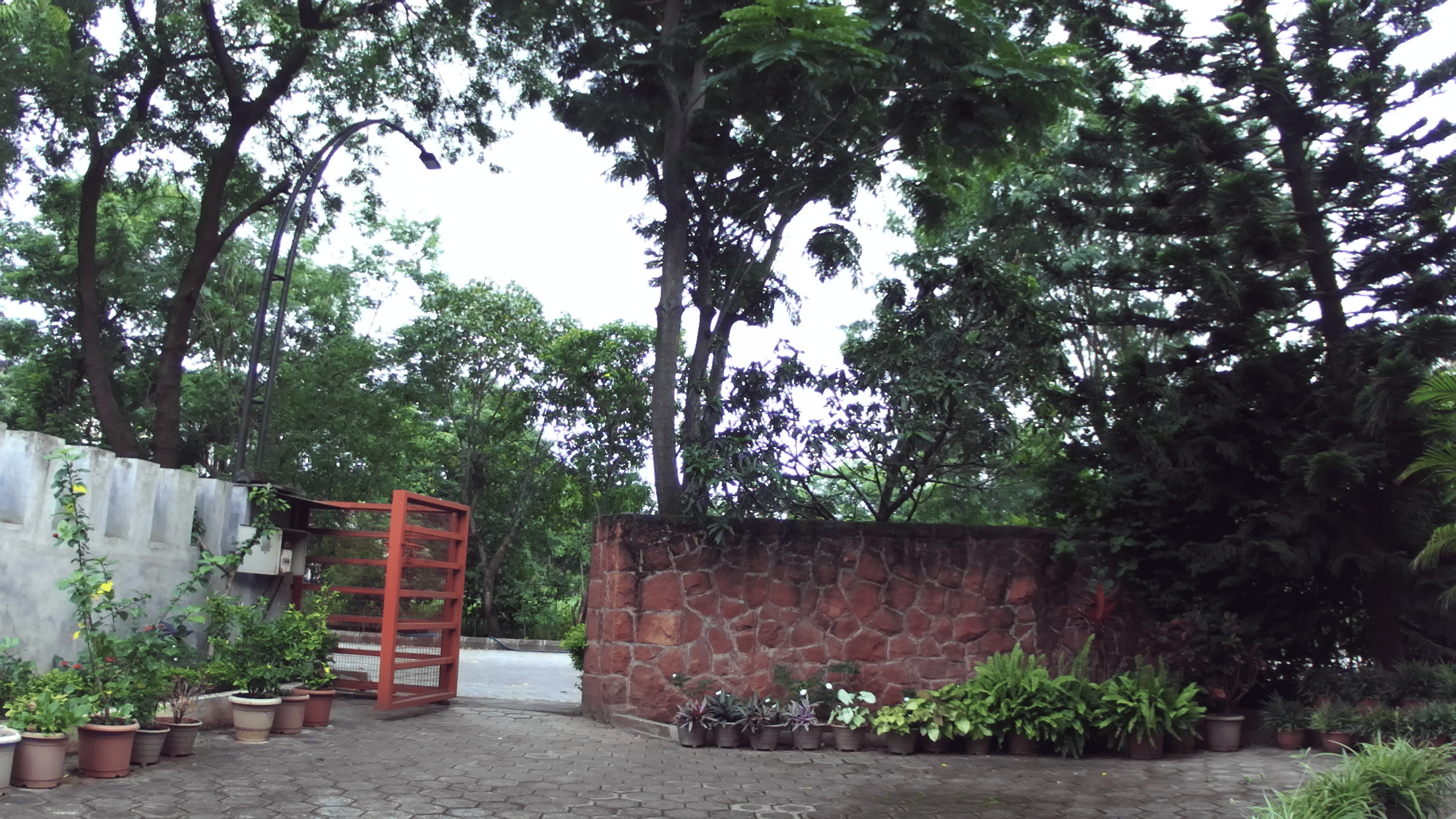}} &
\subfloat[ Gate(right)]{\includegraphics[width = 1.5in]{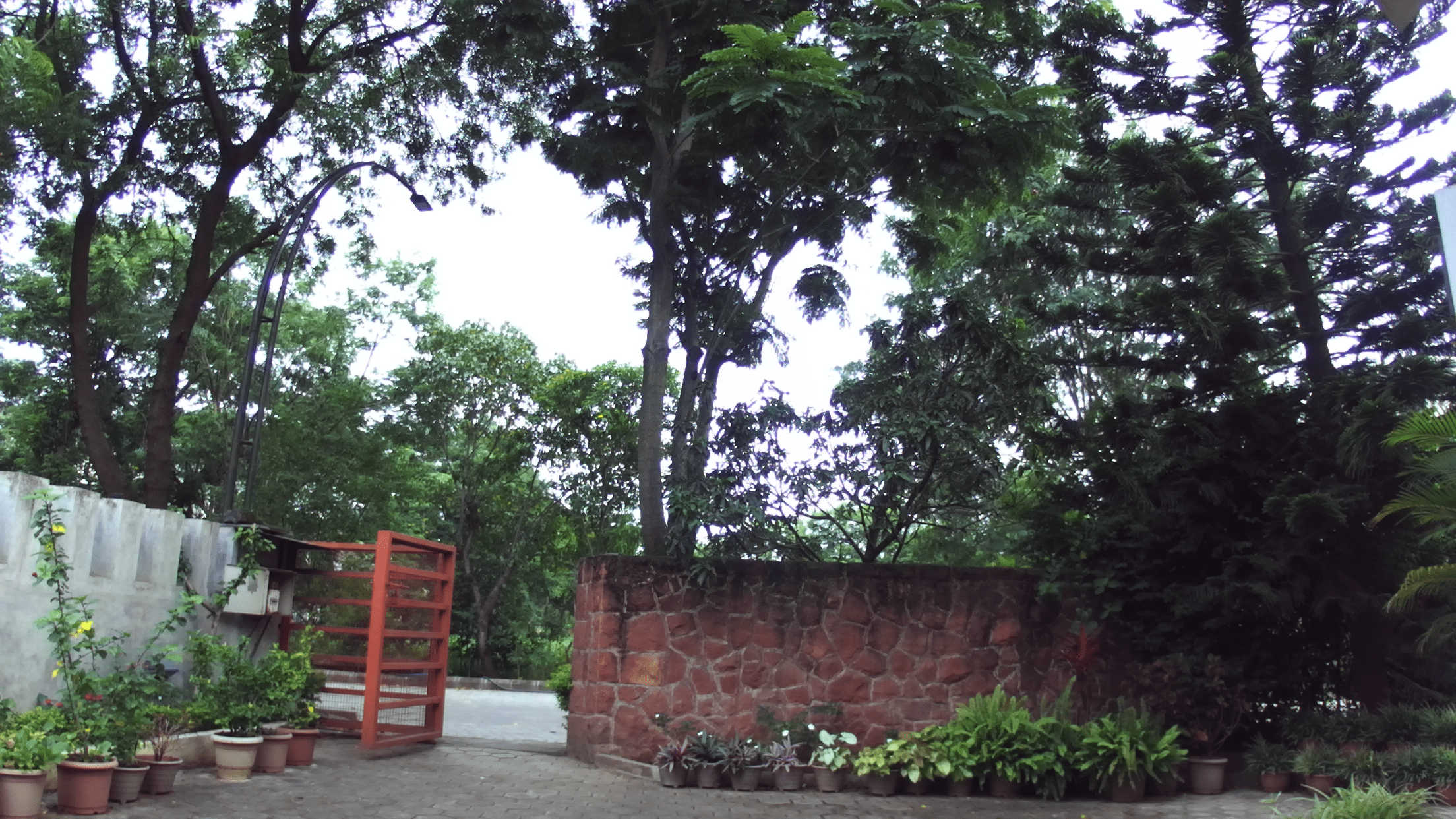}}
\\
\subfloat[ Himalaya Mess(left)]{\includegraphics[width = 1.5in]{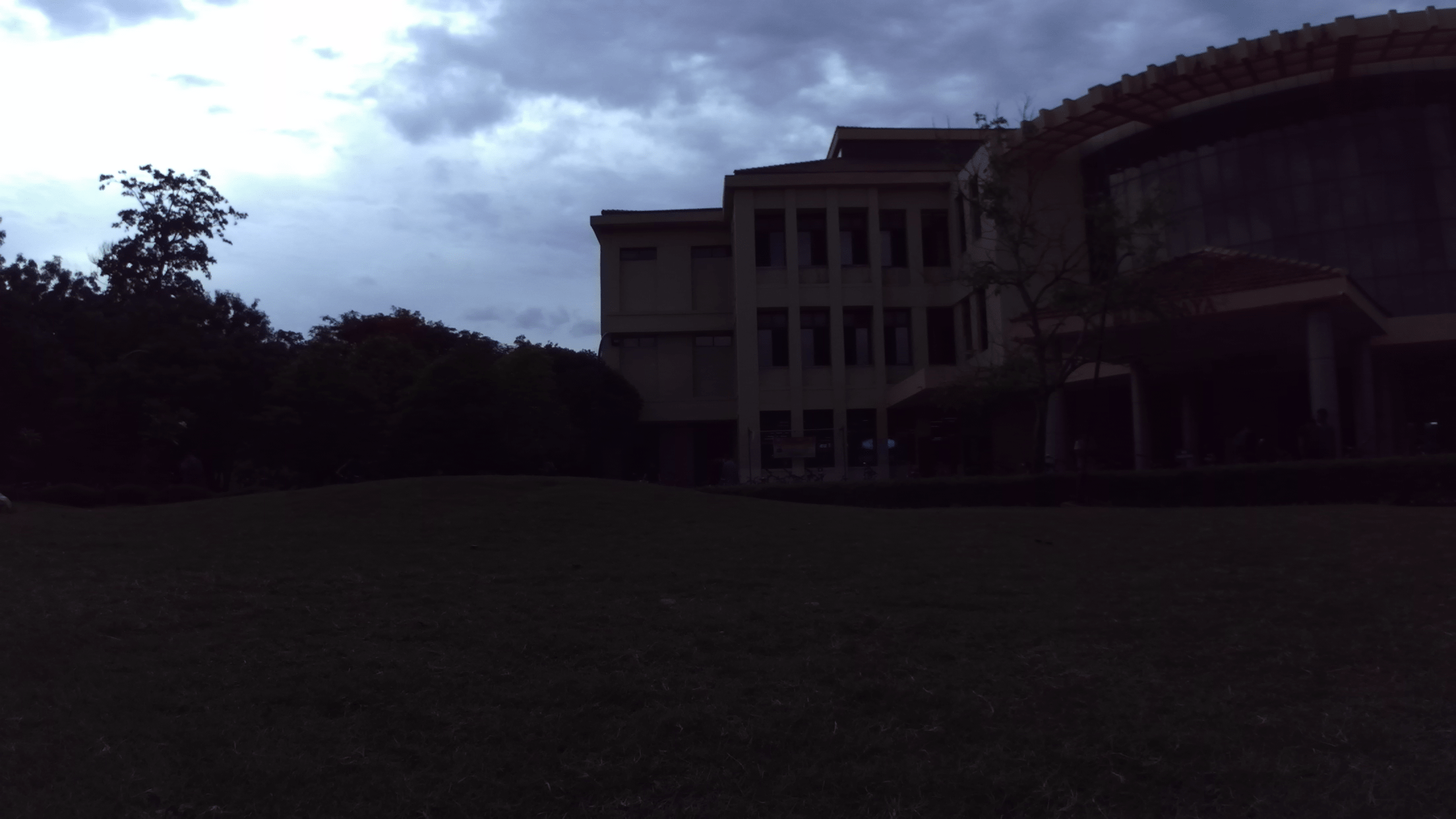}} &
\subfloat[ Himalaya Mess(right)]{\includegraphics[width = 1.5in]{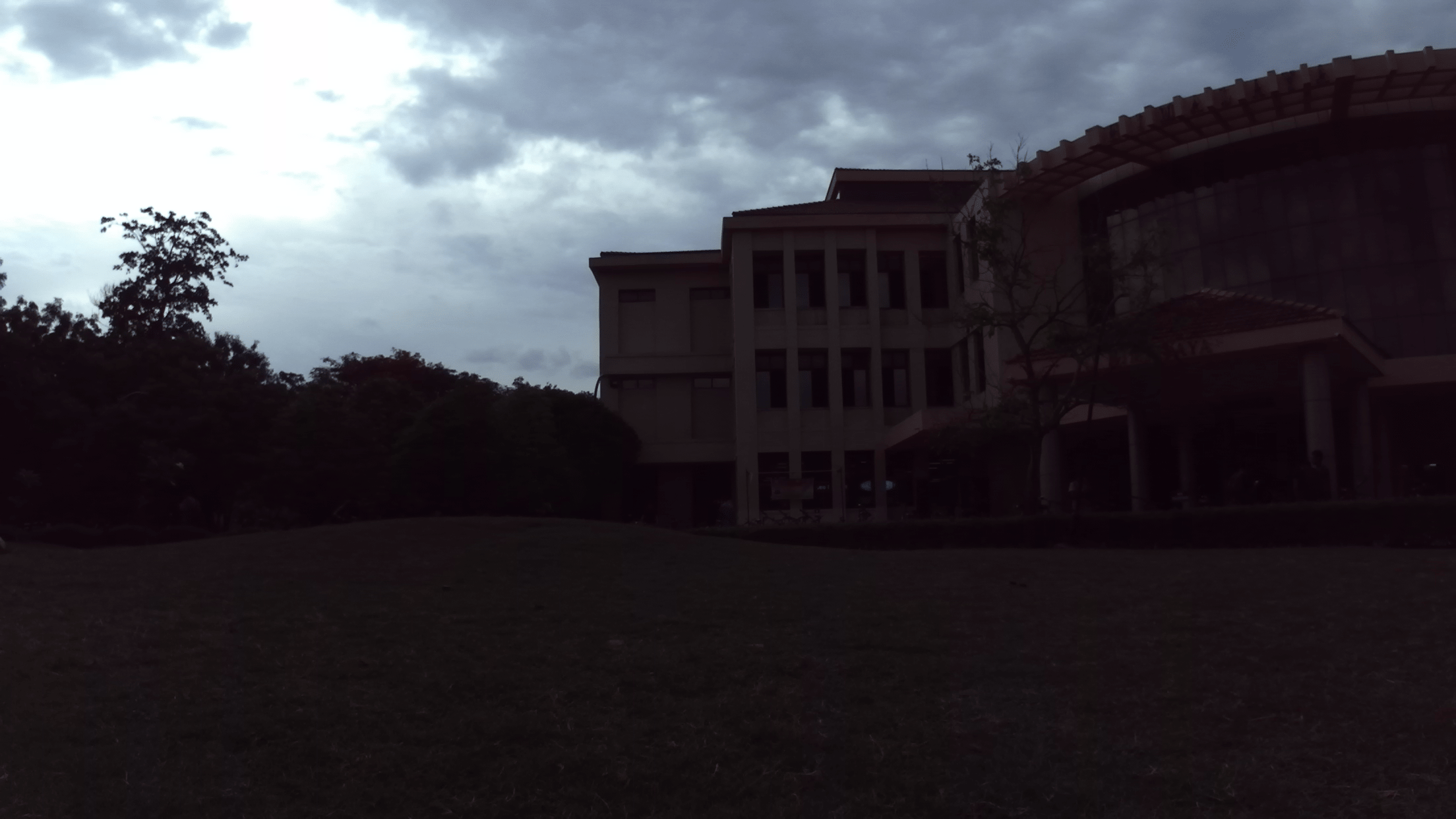}} &
\subfloat[ Himalaya Mess(left)]{\includegraphics[width = 1.5in]{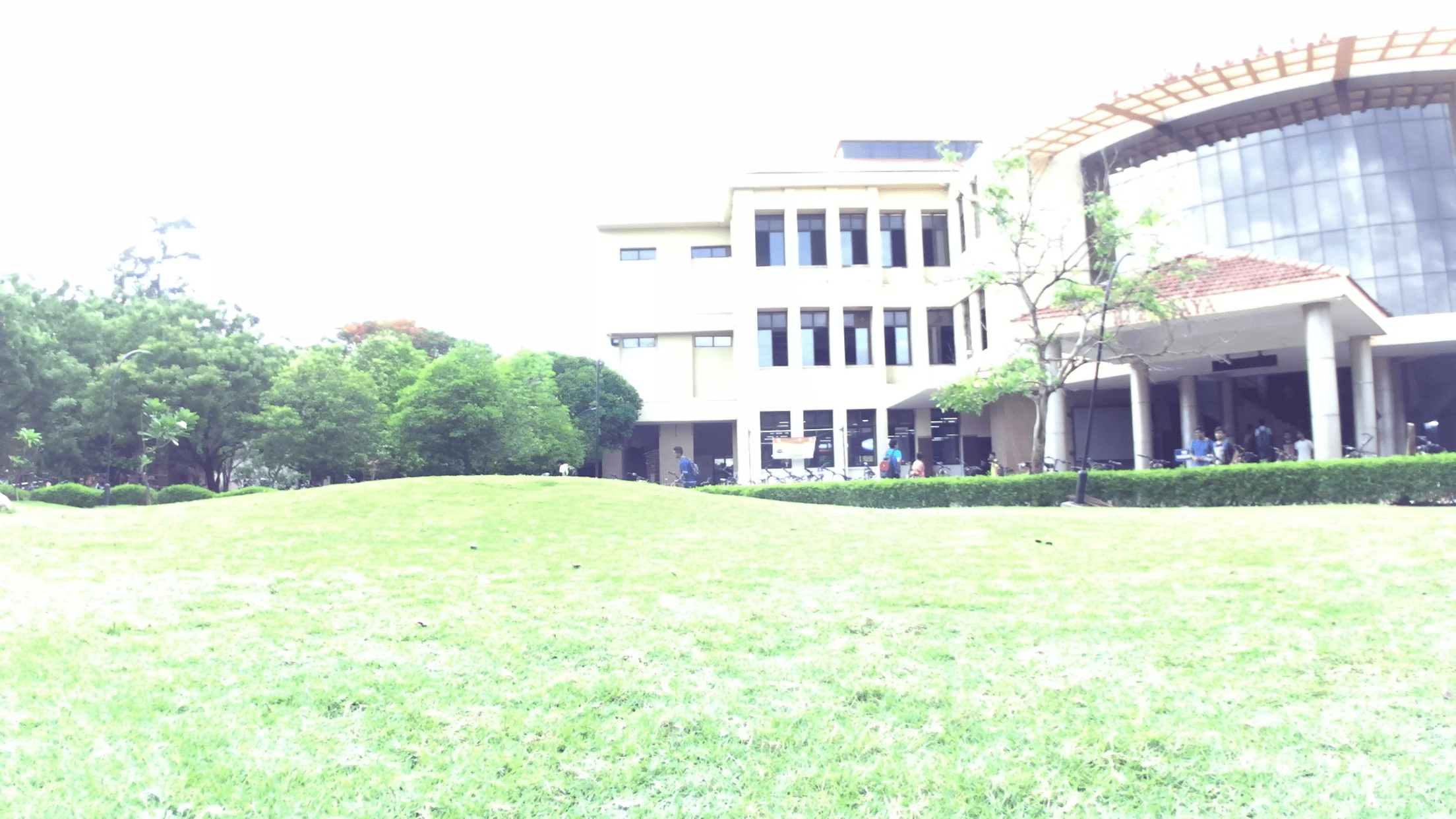}} &
\subfloat[ Himalaya Mess(right)]{\includegraphics[width = 1.5in]{INPUT/Scene3right1-min.png}}
\\
\subfloat[ Room(left)]{\includegraphics[width = 1.5in]{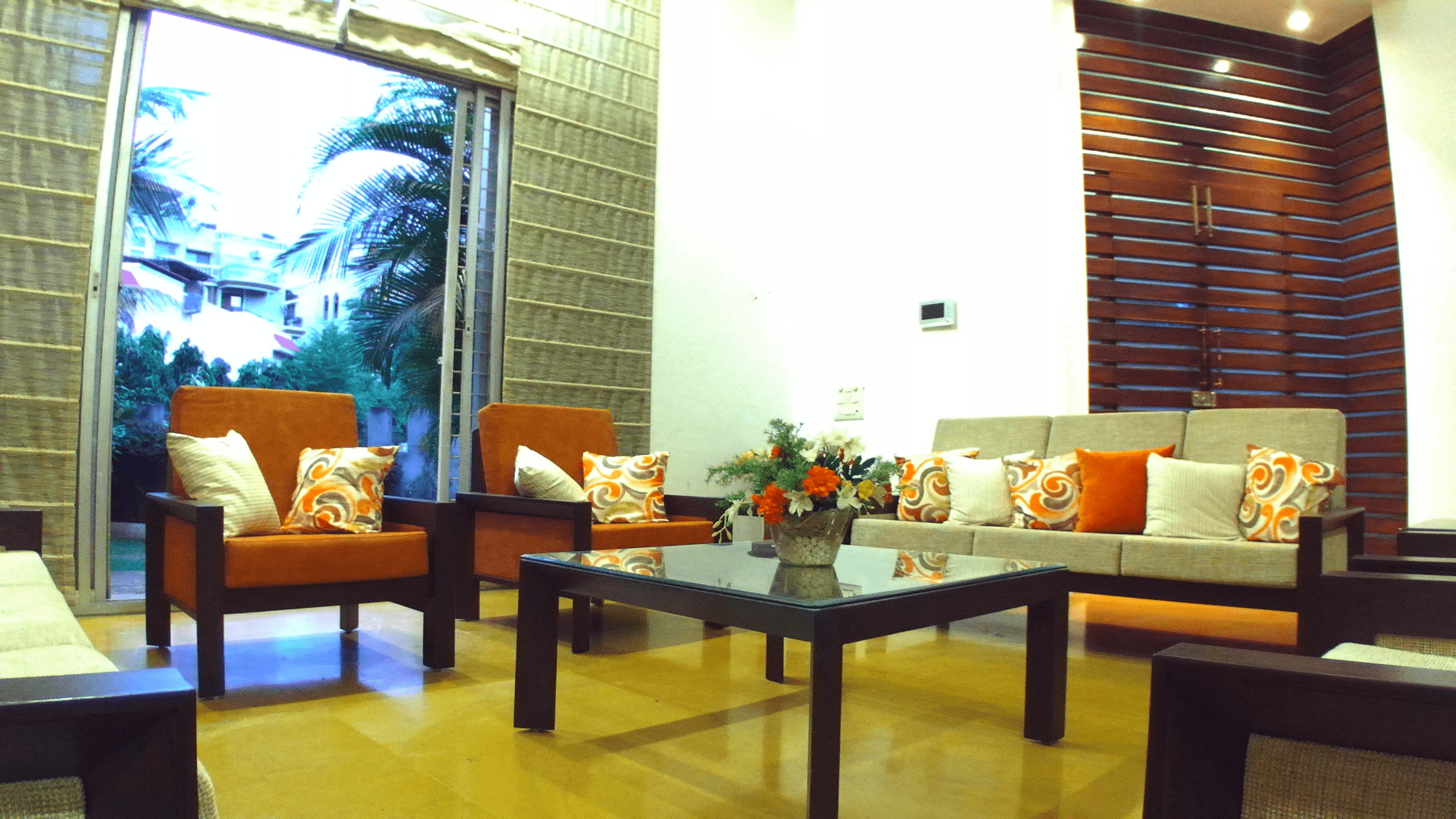}} &
\subfloat[ Room(right)]{\includegraphics[width = 1.5in]{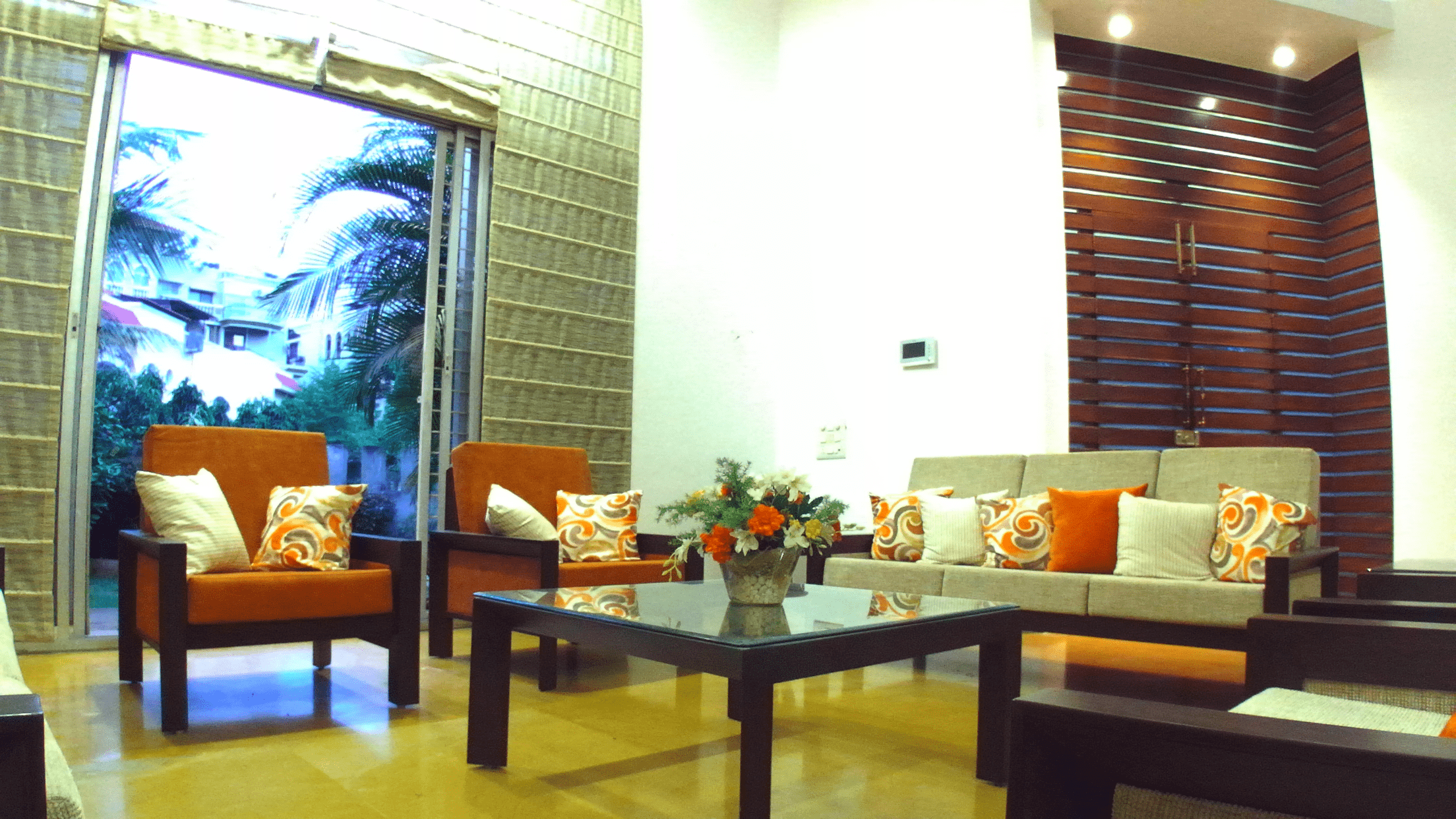}} &
\subfloat[ Room(left)]{\includegraphics[width = 1.5in]{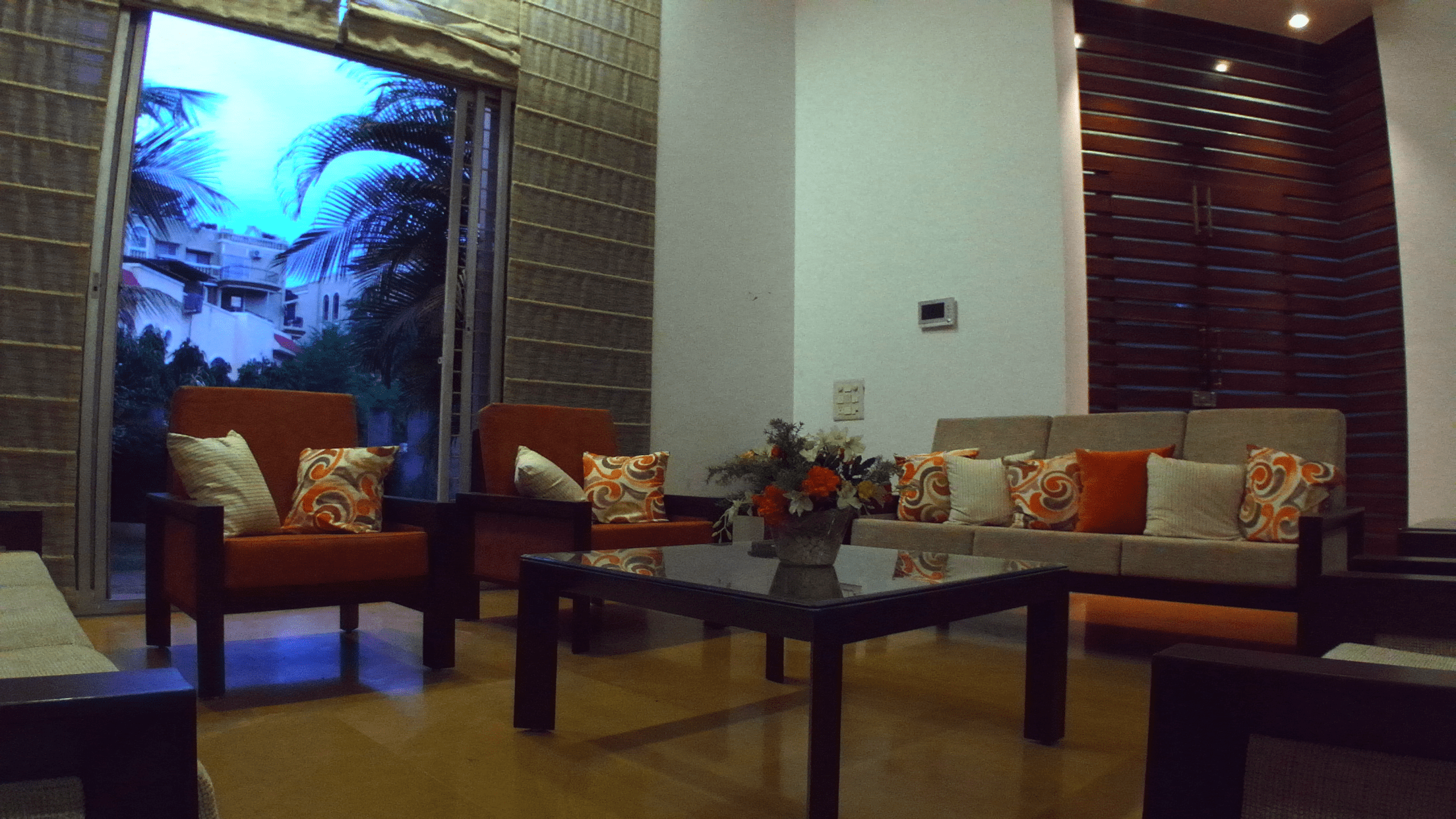}} &
\subfloat[ Room(right)]{\includegraphics[width = 1.5in]{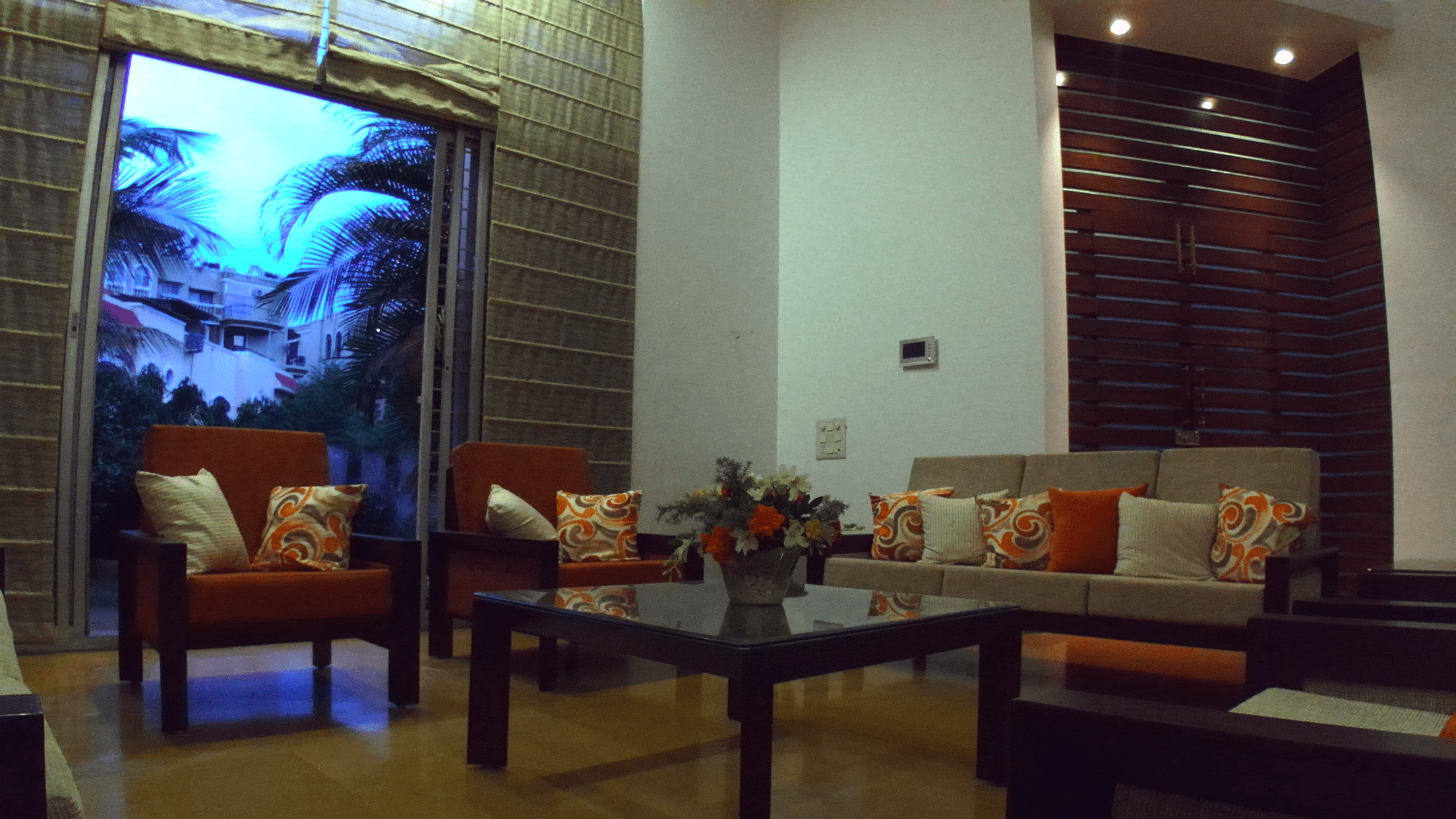}}
\end{tabular}
\caption{Some test multi-exposures stereo images. Note five exposure levels are considered for experiments with every scene.}
\label{Fig2InputScenes}
\end{figure*}

\section{Related Work} 
The non-backward compatible algorithms \cite{RWRef1, RWRef2, RWRef3, RWRef4} are presented for HDR video coding. Motra and Thoma \cite{RWRef2} transform an HDR image into an adaptive Logluv format before compressing by H.264/AVC. Zhang et al. \cite{RWRef3} exploits luminance masking and profile scaling based on tone-mapping curve to enhance the performance of HEVC for HDR video content. 
Mukherjee et al. \cite{RWRef1} converted HDR content into
perceptually uniform color opponent space and encodes the
scene dynamic range using a perceptual transfer function. An error minimization scheme is employed for accurate chroma reproduction. Their rate–distortion characteristics outperforms the existing state-of-the-art coders at bitrates $>=$ 0.4 bits/pixel.
Mantiuk et al. \cite{RWRef4} extends the MPEG-4 encoder. The central component in their approach is inter-frame encoding of HDR video and perception-based HDR luminance-to-integer encoding. Their scheme requires 10-11 bits to encode 12 orders of magnitude of the full perceivable luminance range. Further, the DCT block based coding scheme speed up the global tone mapping operator and separately encode decomposed layers of LDR details and HDR edges to improve synthetic video quality.

The HDR coding methods that support backward compatibility are based on tone mapping models \cite{RWRef5, RWRef6}. Huang et al.~\cite{RWRef5} proposed an HDR compression method based on the matting Laplacian. To preserve HDR local structural details in the tone-mapped LDR image, their scheme treats the HDR image as a guidance image and incorporate matting Laplacian matrix into the objective function. The HDR compression scheme formulated as an optimization problem and extract LDR image with enhanced  details, mitigating severe edge effects or color artifacts. Li et al. \cite{RWRef6} encoded tone mapped LDR image into a JPEG compatible base layer codestream. They applied inverse tone mapping operation to approximate the HDR image from the reconstructed LDR image. The residual generated between original HDR image 
and approximated HDR image is further encoded. The progressive JPEG-based coding of extension layer adaptively set different quantization parameters for different blocks, weighing more for blocks with higher saliency, and encoded them first. The coding quality is adapted to the visual saliency region. Watanabe et al. \cite{RWRef7} presented another backward compatible scheme to the legacy JPEG standard. Their two layer coding method using the histogram packing technique achieves better lossless compression performance than state-of-the-art JPEG-XT \cite{Ref2}. Mai et al. \cite{RWRef8} addresses the issue to find the tone-curve that minimizes mean square error in the reconstructed HDR sequence by formulating the problem as a numerical optimization. Unlike existing lossless HDR coding strategies, Feng et al. \cite{RWRef9} approach does not require mapping of HDR into LDR images before conventional coding tools can be applied. Their proposed mapping-free post HDR image compression scheme excludes integer to floating-point conversion and tone-mapping. The scheme is applicable to compress RGBE and OpenEXR HDR contents. 

\section{Proposed Multi-exposure Stereo Coding Scheme} 
A workflow of the proposed coding scheme for multi-exposure stereo images is illustrated in Fig.~\ref{Fig1}. The detail of each component in the compression pipeline is described in this section. 

\subsection{Color Space Transform} 
We performed the color space transform of input multi-exposure stereo images to commonly used $Y^{'}C_bC_r$ \cite{YCbCr1, IPT4} and recently introduced IPT color opponent space \cite{IPT1, IPT2, IPT3, IPT4}.

The $YC_bC_r$ color space has standardized within the recommendation of ITU-R BT.601 \cite{YCbCr2}. The $Y$ is luma component and $C_b$ and $C_r$ are the blue-difference and red-difference chroma components. The $Y^{'}$ (with prime) denotes the luminance. $Y^{'}$ is distinguished from $Y$, meaning that light intensity is nonlinearly encoded based on gamma corrected RGB primaries. Applying Gamma correction to the luminance channel reduces the perception of quantization error. This representation is commonly referred as $Y^{'}C_bC_r$.

Non-Constant Luminance $Y^{'} C_b C_r$ is commonly used color space for the distribution of SDR signals. NCL $Y^{'} C_b C_r$ color distribution format has some limitations. It cannot fully de-correlate intensity information from chroma information since it is constrained by RGB color primaries which keep varying. In $Y^{'} C_b C_r$, color difference weights can be obtained by filling a color volume and not based on perceptual model. Constant luminance $Y^{'} C_b C_r$ was added in ITU-R BT.2020 to regulate $Y^{'} C_b C_r$ in de-correlating intensity from chroma effectively. 

The perceptually uniform IPT color opponent space is proposed by Ebner and Fairchild \cite{IPT1}. The I, P, and T coordinates in IPT color space, represent the lightness, the red-green and the yellow-blue dimensions, respectively \cite{IPT5}. The IPT model does not detrimentally affects other color appearance attributes, while predicting the hue components. It mitigates the hue compressibility issues and maintains the perceptual uniformity of CIELAB/LUV \cite{IPT2}. Therefore, ITP color space is employed frequently in latest HDR/WCG video compression schemes as a new signal format \cite{RWRef1, IPT3} compared with NCL $Y^{'} C_b C_r$. Since IPT color space is derived using perceptual mechanisms of the Human Visual System (HVS). It is determined to better de-correlate intensity and chroma information, while simultaneously minimizing the perceptual distortion. 

The analysis and simulation results presented by Lu et al. \cite{IPT3} characterize IPT as an improved intensity prediction (constant luminance) that prevents visible quantization artifacts at a given bit depth. They validate IPT properties to de-correlate the chroma from luma components over $YC_bC_r-PQ$ used for HDR signals. The $YC_bC_r-PQ$ applied transfer function in linear RGB space. The subjective and quantitative experiments found IPT is better in prediction lines of constant hue for increasing values of chroma and constant luminance encoding more effective in chroma downsampling, such as 4:2:0. The psychophysical experiments conducted by \cite{IPT1, IPT2, IPT3, IPT5} validate the advantages of ITP over $YC_bC_r-PQ$ for compression efficiency in HDR and wide-gamut imaging. Zerman et al. \cite{IPT4} demonstrated comparable performance of IPT with $Y^{'}C_b C_r$ in HDR video compression. However, Zerman et al. \cite{IPT4} found no evidence of the ITP color space being significantly better than $Y^{'} C_b C_r$ for HDR video compression. Their experiments revealed that the choice of color space influence compression performance for HDR video, in general, little.

Therefore, in our proposed coding scheme, we considered both $Y^{'} C_b C_r$ and IPT signal format in testing coding performance of stereo multi-exposure images. The constant-luminance $Y^{'} C_b C_r$ representation commonly can decorrelate the luminance and chrominance information, especially in the present SDR scenario, \textit{i.e.}, coding stereo multi-exposure images formulation. Further, IPT is chosen as one of the color formats because of its desirable properties in color quantization, constant luminance, hue characteristic, chroma subsampling.

\begin{table}[t]
    \centering
    \caption{Experimental Parameters}
    \begin{tabular}{|c|c|c|c|c|}
    \hline
    \multicolumn{1}{|c|}{Platform} & \multicolumn{4}{c|}{3D-HTM 16, JPEG XT} \\
    \hline
    \multicolumn{1}{|c|}{QP} & \multicolumn{4}{c|}{5,10,15,20} \\
    \hline
    DATA & GATE & GARDEN & HIMALAYA MESS & ROOM\\
    \hline
     \multicolumn{1}{|c|}{Resolution} & \multicolumn{4}{c|}{2208 $\times$ 1242    } \\
    \hline
    \end{tabular}
    \label{ExperimentalPar1}
\end{table}

\begin{figure*}[t] 
\centering

\begin{tabular}{cccc}
\subfloat[ Garden]{\includegraphics[width = 3.0in]{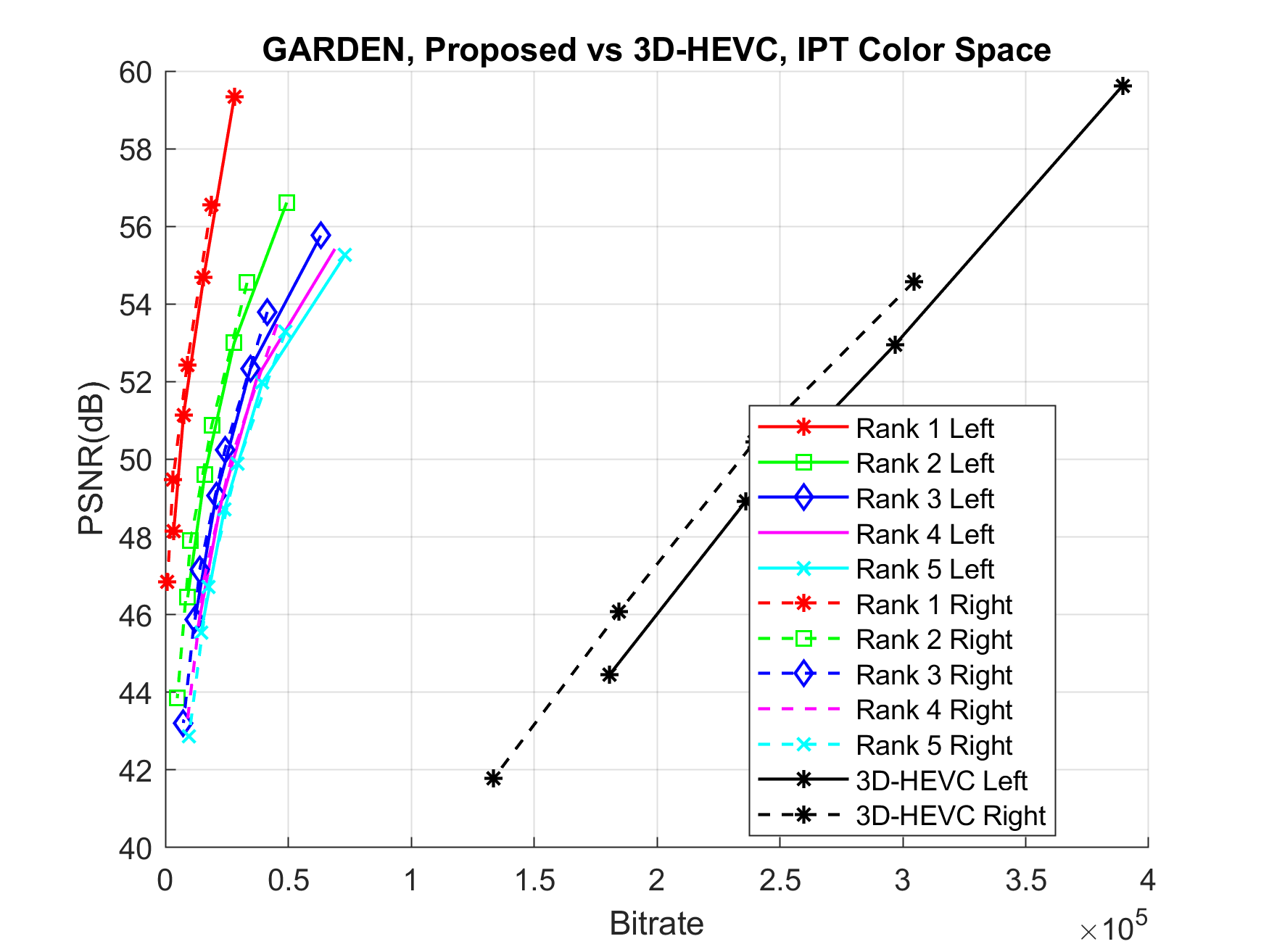}} &
\subfloat[Gate]{\includegraphics[width = 3.0in]{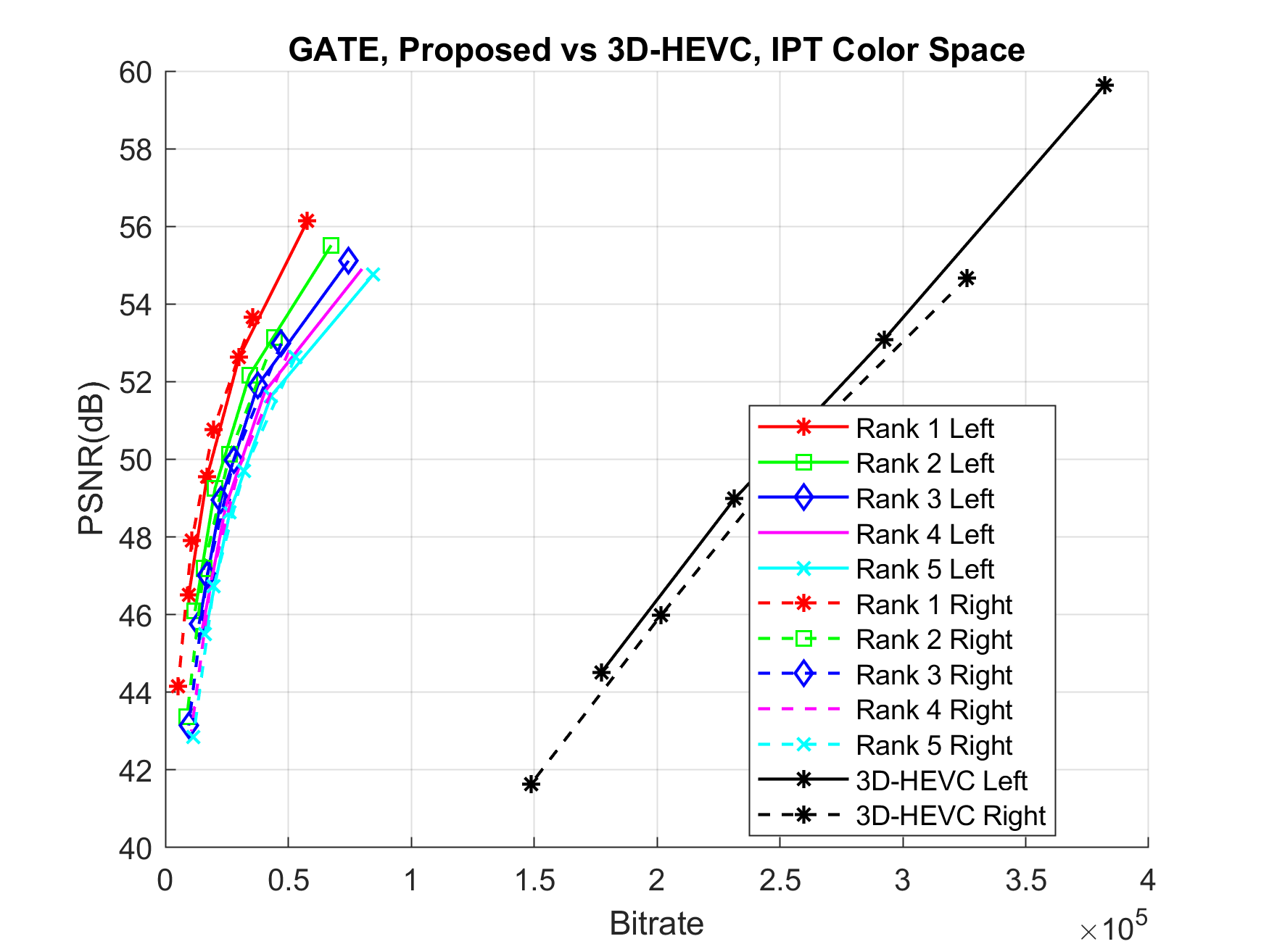}} 
 \\
\subfloat[ Himalaya Mess]{\includegraphics[width = 3.0in]{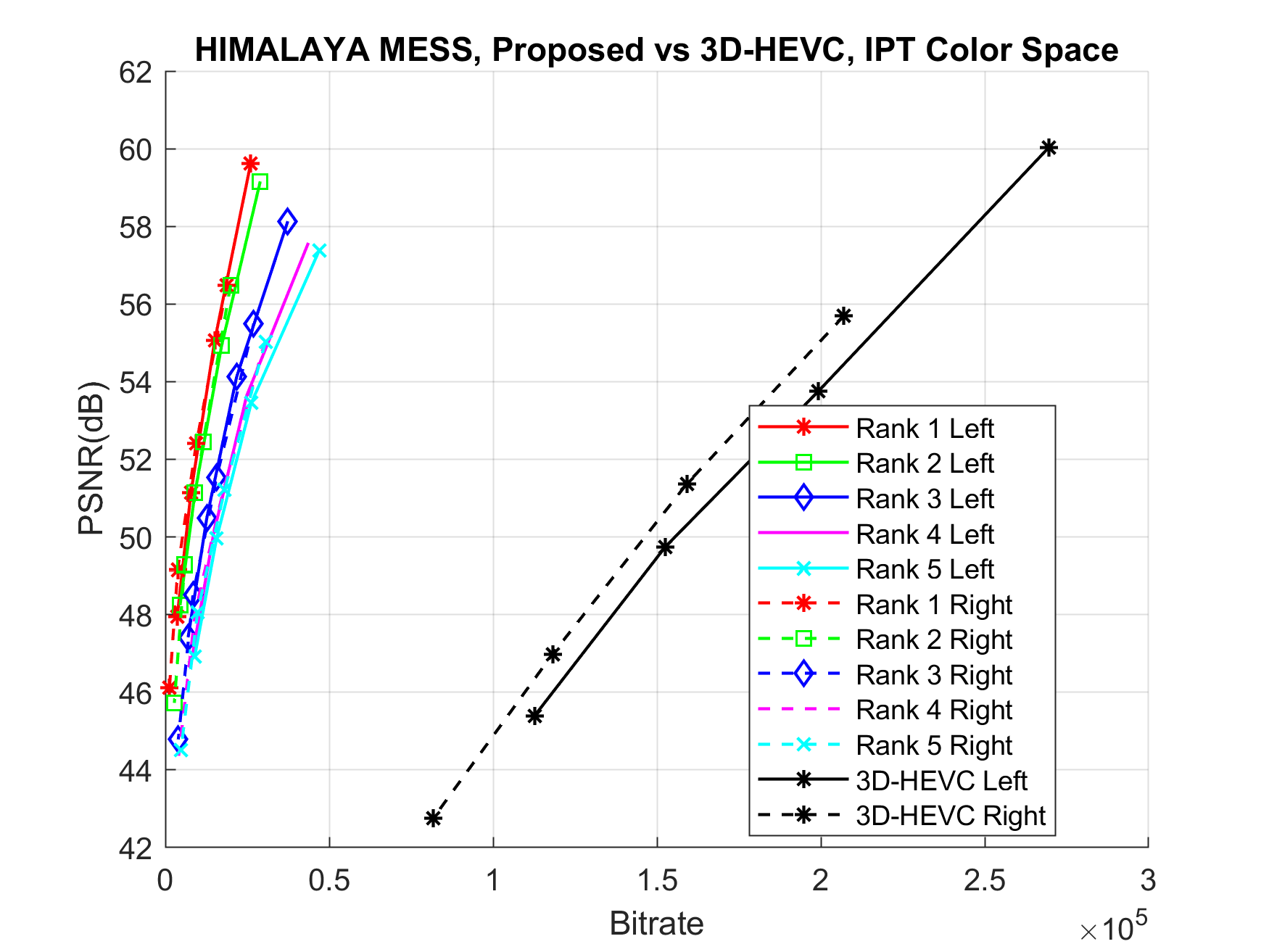}} &
\subfloat[ Room]{\includegraphics[width = 3.0in]{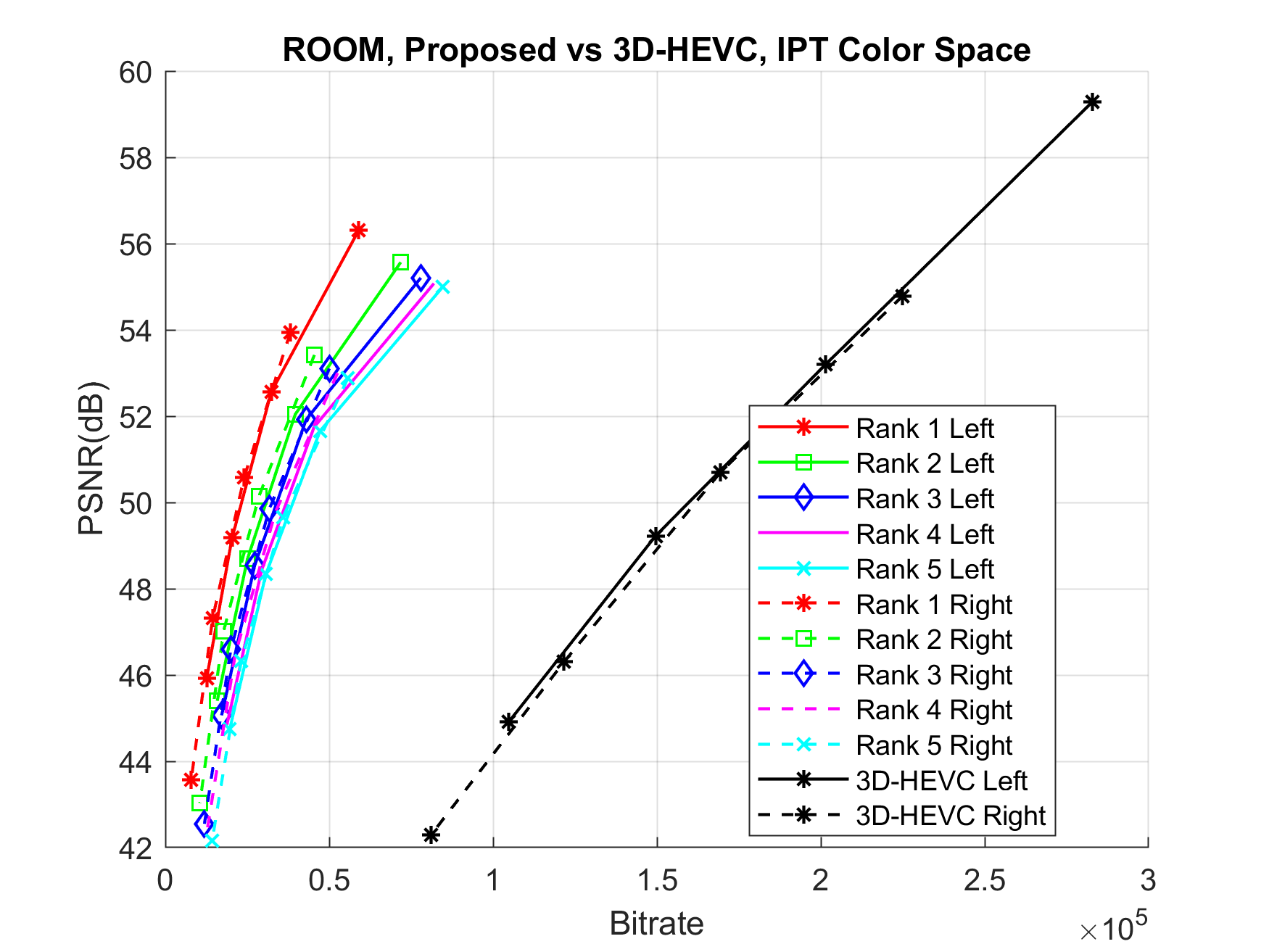}} 
\end{tabular}
\caption{Comparative analysis of proposed scheme at varying tensor rank approximation with 3D-HEVC (IPT Color Space).}
\label{Ourvs3DHEVCIPT}
\end{figure*}
\subsection{Tucker ALS Low Rank Approximation} 
We adopted an alternating least square (ALS) approach for computing a low-rank Tucker decomposition of stack of multi-exposure stereo images in IPT or $Y^{'}$ $C_b$ $C_r$ color space \cite{Ref10}. Let us denote tensor created by stacking multi-exposure stereo images as $\mathfrak{T}_{MES}^{c}$. The tensor $\mathfrak{T}_{MES}^{c}$ is efficiently approximated by a core tensor contracted by orthogonal matrices along each mode. Suppose $\mathfrak{T}_{MES}^{c} \in \textrm{R}^{s_1 \times ... \times s_N}$ is an order $N$ tensor with dimensions $s_1 \times ....\times s_N$.
The Tucker decomposition of $\mathfrak{T}_{MES}^{c}$ is computed by

\begin{multline} 
\mathfrak{T}_{MES}^{c} \approx [[ \mathfrak{Q}; \textbf{S}^{(1)}, \textbf{S}^{(2)} ..., \textbf{S}^{(N)} ]] \\
= \mathfrak{Q} \times_1 \textbf{S}^{(1)} \times_2 \textbf{S}^{(2)} ... \times_N \textbf{S}^{(N)}
\label{Exp1}
\end{multline}
Rewriting the expression (\ref{Exp1}) element-wise gives
\begin{multline} 
\mathfrak{T}^{a}_{LF}(t_1,...., t_N) \approx \sum_{ \{q_1,....,q_N \}}  \mathfrak{Q}(q_1,....,q_N) \\ \prod_{r \in \{ 1,...., N \} } \textbf{S}^{(r)}(t_r,q_r)
\end{multline} 
where, $\mathfrak{Q}$ is a core tensor of coefficients. The core tensor $\mathfrak{Q}$ is of order $N$ with Tucker ranks (or dimensions) $R_{1} \times ... \times R_N$, and the matrices $\textbf{S}^{(r)}(t_r,q_r) \in R^{s_n \times R_n}$ have orthonormal columns. Let we assume, without loss of generality, each $R_n = R$ for $n \in \{1,...,N \}$.
\\
\subsubsection{Initialize Tucker-ALS algorithm}
The higher-order singular value decomposition (HOSVD) is employed to initialize $ \mathfrak{T}_{MES}^{c} \in \textrm{R}^{s_1 \times ... \times s_N}$ by adopting a truncation based scheme \cite{Ref7}. The truncated HOSVD is computationally efficient and provides an appropriate starting point for the Tucker-ALS algorithm.
Let $\mathfrak{T}_{MES}^{c}$ admits a higher order singular value decomposition 
\begin{equation} 
\mathfrak{T}_{MES}^{c} = (\textbf{S}^{(1)}, \textbf{S}^{(2)},...,\textbf{S}^{(N)}) \hspace{2pt} \cdot \hspace{2pt} \mathfrak{Q}
\end{equation} 
where, the factor matrix $\textbf{S}^{(r)}$ is an orthogonal $s_{r} \times s_{r}$ matrix, obtained from the SVD of mode-$r$ unfolding of $\mathfrak{Q}$,
\begin{equation} 
\textbf{S}^{(r)} = U_{r} \Sigma_{r} V_{r}^{T},
\end{equation} 
and the core tensor $ \mathfrak{Q} \in \textrm{R}^{s_1 \times ... \times s_N}$
can be recovered from
\begin{equation} 
\mathfrak{Q} = (\textbf{S}^{(1)^T}, \textbf{S}^{(2)^T},...,\textbf{S}^{(N)^T}) \hspace{2pt} \cdot \hspace{2pt} \mathfrak{T}_{MES}^{c} 
\end{equation} 
A low multi-linear rank approximation is constructed to approximate $\mathfrak{T}_{MES}^{c}$ by a rank-$(k_1; k_2; ... ; k_N)$ tensor
$\mathfrak{\hat{T}}_{MES}^{c}$, where $k_r \leq s_r$, $\forall$ $1 \leq r \leq N $. 
The truncated HOSVD factor matrix $\hat{U}_{r}$ is obtained from a truncated SVD of the mode-$r$ unfolding of the tensor, such as
\begin{equation} 
\textbf{S}^{(r)} = U_{r} \Sigma_{r} V_{r}^{T} = [\bar{U}_{r} \hspace{5pt} \hat{U}_{r}] \begin{bmatrix}
\bar{\Sigma}_{r}  & \\ 
 & \hat{\Sigma}_{r}
\end{bmatrix} \begin{bmatrix}
\bar{V}^{T}_{r} \\ \vspace{5pt}
\hat{V}^{T}_{r}
\end{bmatrix}
\end{equation} 
with $\bar{U}_{r} \in R^{s_r \times k_r}$. The approximation $\mathfrak{\hat{T}}_{MES}^{c}$ of $\mathfrak{T}_{MES}^{c}$  
is obtained considering an orthogonal projection onto the tensor basis 
\begin{multline} 
\mathfrak{\hat{T}}_{MES}^{c} := \varpi_1 \varpi_2 \cdot \cdot \cdot \varpi_N  \mathfrak{T}_{MES}^{c} := (\bar{U}_{1}\bar{U}^{T}_{1},  ..., \bar{U}_{N}\bar{U}^{T}_{N}) \hspace{2pt} \cdot \hspace{2pt} \\ \mathfrak{T}_{MES}^{c} 
=: (\bar{U}_{1},  ..., \bar{U}_{N}) \hspace{2pt} \cdot \hspace{2pt} \mathfrak{\bar{Q}} \approx \mathfrak{T}_{MES}^{c}
\end{multline} 
where, $(\bar{U}^{T}_{1},  ..., \bar{U}^{T}_{N}) \hspace{2pt} \cdot \hspace{2pt} \mathfrak{T}_{MES}^{c} =: \mathfrak{\bar{Q}} \in \textrm{R}^{s_1 \times ... \times s_N}$ is defined as truncated core tensor. The computed $\mathfrak{\hat{T}}_{MES}^{c}$ be the rank-$(k_1; k_2; ... ; k_N)$ T-HOSVD of   
$\mathfrak{T}_{MES}^{c}$ and used to initialize the Tucker-ALS algorithm. One 
can also use interlaced HOSVD for more efficient computation of T-HOSVD of 
$\mathfrak{T}_{MES}^{c}$ \cite{Ref10}.

The conventional ALS method for Tucker decomposition adopted a higher-order orthogonal iteration (HOOI) procedure to fix all except one factor matrix, and then computes a low-rank matrix factorization to update that factor matrix and the core tensor \cite{Ref8, Ref9}. The conventional Tucker-ALS approach factorizes into a product of an orthogonal matrix $\textbf{S}^{(n)}$ and the core tensor $\mathfrak{Q}$ to update the $n$-th factor matrix as
\begin{multline} 
\textsl{Y}^{(n)} = \mathfrak{T}_{MES}^{c} \times_1  \textbf{S}^{(1)^T} \hspace{3pt} \cdot \cdot \cdot \hspace{3pt} \times_{n-1} \textbf{S}^{(n-1)^T} \times_{n+1} \textbf{S}^{(n+1)^T} \hspace{3pt} \\ \cdot \cdot \cdot \hspace{3pt}
\times_{N} \textbf{S}^{(N)^T}
\end{multline}
and $\textbf{Y}_{n}^{n} \approx \textbf{S}^{(n)} \textbf{Q}_{(n)}$. This factorization can be performed by considering $\textbf{S}^{(n)}$ to take $R^{n}$ leading left singular vectors of $\textbf{Y}_{n}^{(n)}$. We computed singular vectors by finding the left eigenvectors of the Gram matrix $\textbf{Y}_{n}^{(n)}\textbf{Y}_{n}^{(n)^{T}}$ as suggested by Linjian and Edgar \cite{Ref10}. This helps to avoid large SVD and maintain consistency of the singular vectors signatures across ALS sweeps.

Further, we adopted a pairwise perturbation scheme to accelerate the ALS procedure as implemented by Linjian and Edgar \cite{Ref10}. We computed the pairwise perturbation operators, which correlate a pair of factor matrices. Then, the quadratic subproblems formed are repeatedly updated for each tensor using tensors formed from such factor matrices. In proposed formulation, there is not much variation in the factor matrices, so updates are reasonably accurate. The approximation is derived to asymptotically reduce the computational complexity.

We approximate $\tilde{\textsl{Y}}^{(n)} \approx \textsl{Y}^{(n)}$ following 
similar pairwise perturbation algorithm for Tucker-ALS derived by Linjian and Edgar \cite{Ref10}. Let $\textsl{Y}^{(n)}$ be expressed as 
\begin{equation} 
\textsl{Y}^{(n)} = \mathfrak{T}_{MES}^{c} \hspace{5pt} \times_{i=1,i \neq n}^{N}
\hspace{5pt} (\textbf{S}_{p}^{(i)^{T}} + d\textbf{S}^{(i)^{T}})
\label{Eq9}
\end{equation} 
Note $\textbf{S}_{p}^{(n)}$ denote the $\textbf{S}^{(n)}$ computed with a standard ALS step at some (\textit{i.e.}, $p$) number of steps preceding to the present one.
Thus, $\textbf{S}^{(n)}$ at the current step can be expressed as
\begin{equation} 
\textbf{S}^{(n)} = \textbf{S}_{p}^{(n)} + d\textbf{S}^{(n)}
\end{equation} 
The expression (\ref{Eq9}) can be rewritten with pairwise perturbation considering 
$\textsl{Y}^{(n)}$ contracted with $\textbf{S}_{p}^{(n)}$ and first order terms in 
$d\textbf{S}^{(n)}$ 
\begin{equation} 
\tilde{\textsl{Y}}^{(n)} = \textsl{Y}_{p}^{(n)} + \sum_{i=1,i \neq n}^{N} \textsl{Y}_{i,p}^{(n)} \times_i  d\textbf{S}^{(i)^{T}},
\end{equation} 
where, 
\begin{equation} 
 \textsl{Y}_{p}^{(n)} = \mathfrak{T}_{MES}^{c}  \times_{l=1,l \neq n}^{N} \textbf{S}_{p}^{(l)^{T}}
\end{equation} 
and 
\begin{equation} 
\textsl{Y}_{p}^{(i,n)} = \mathfrak{T}_{MES}^{c} \times_{j \in \{1,...,N \} \setminus  \{i,n\} } \textbf{S}_{p}^{(j)^{T}}
\end{equation} 
Given $\textsl{Y}_{p}^{(n)}$ and $\textsl{Y}_{p}^{(i,n)}$, the $\tilde{\textsl{Y}}^{(n)}$ is computed for all $n \in \{1,...,N\}$ efficiently. We adopted dimension trees based approach for the computation of the pairwise perturbation operators $\textsl{Y}_{p}^{(n)}$ and $\textsl{Y}_{p}^{(i,n)}$.

\begin{figure*}[t] 
    \centering

\begin{tabular}{cccc}
\subfloat[ Garden]{\includegraphics[width = 3.0in]{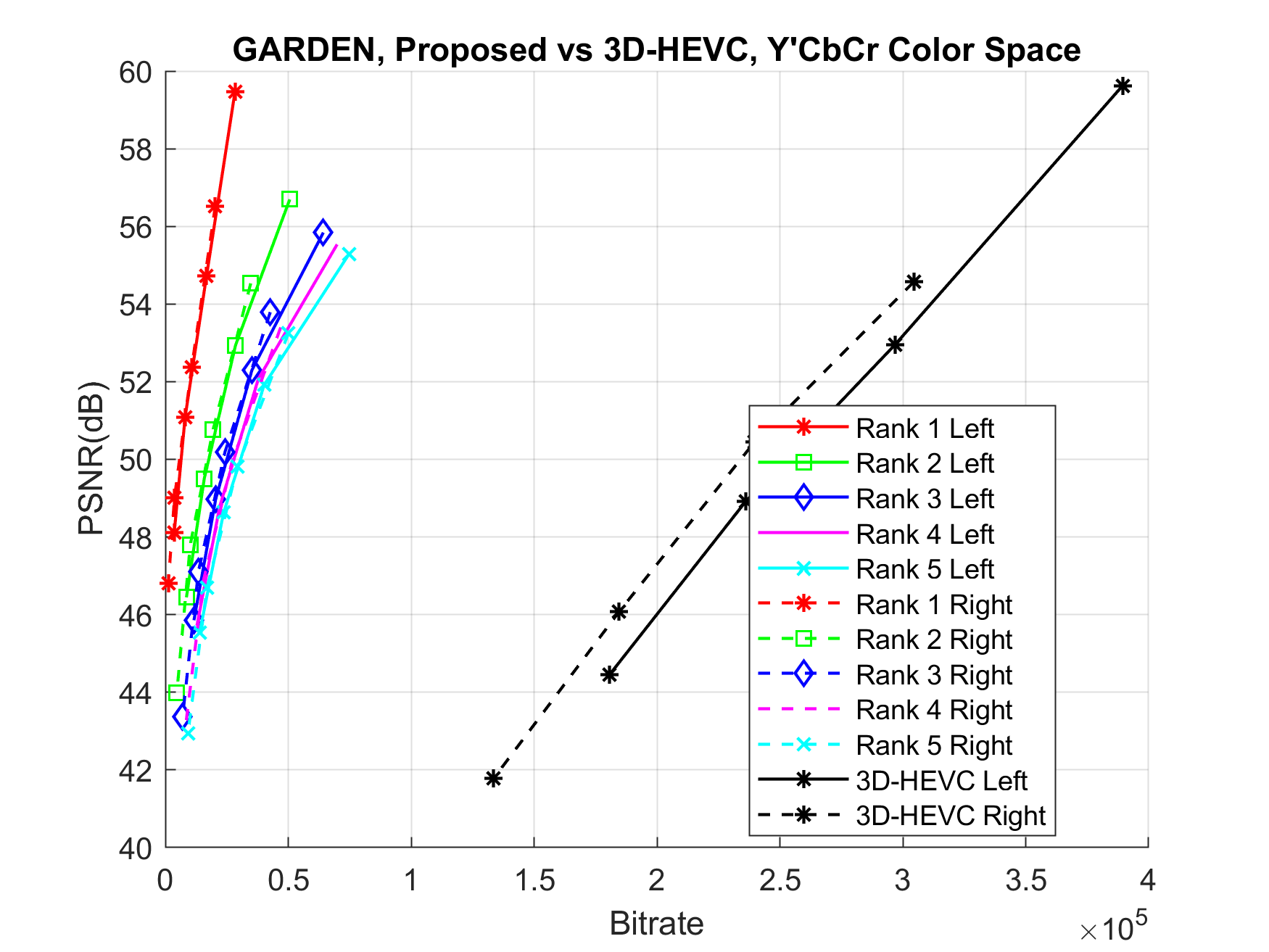}} &
\subfloat[Gate]{\includegraphics[width = 3.0in]{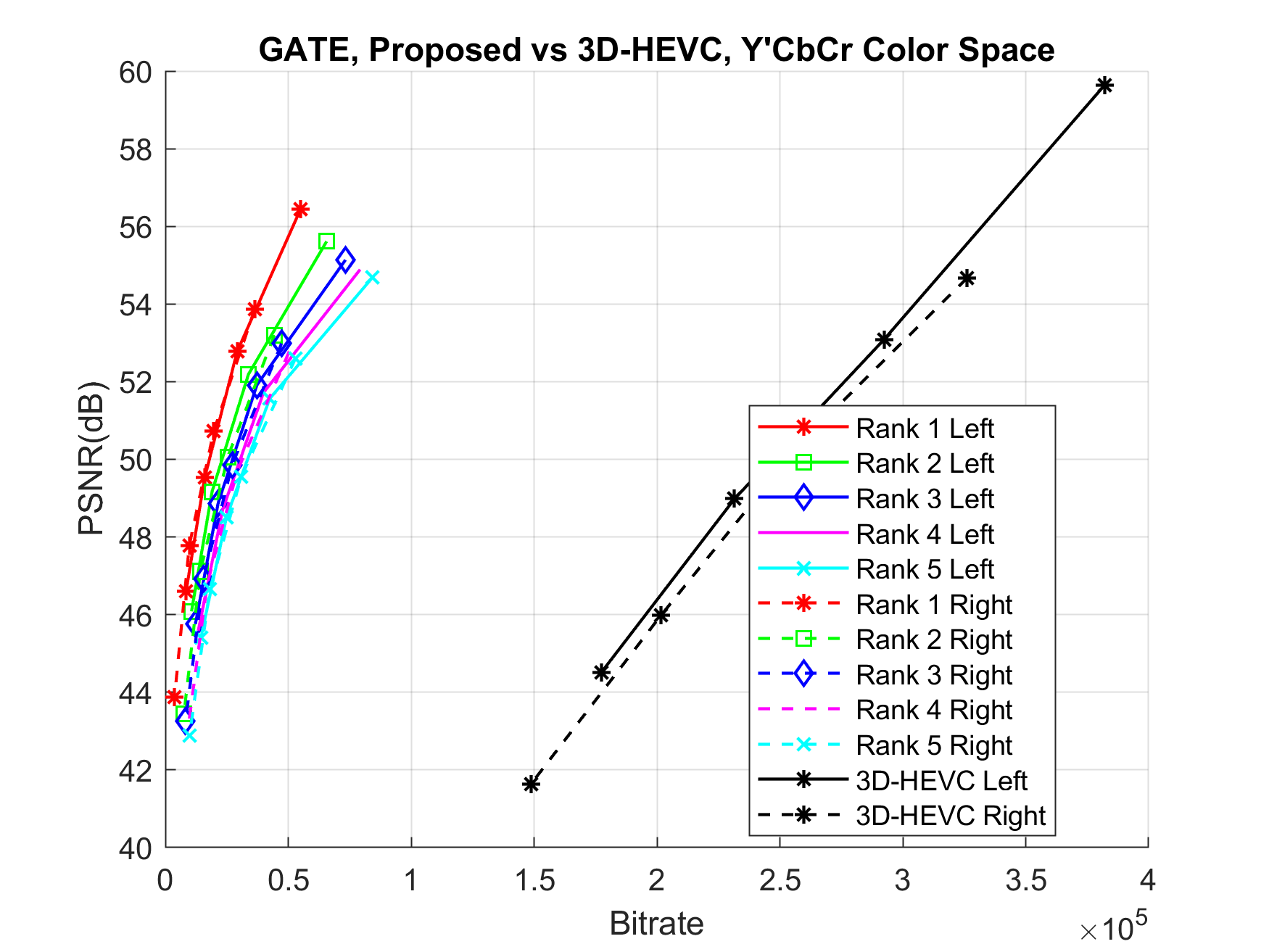}} 
 \\
\subfloat[ Himalaya Mess]{\includegraphics[width = 3.0in]{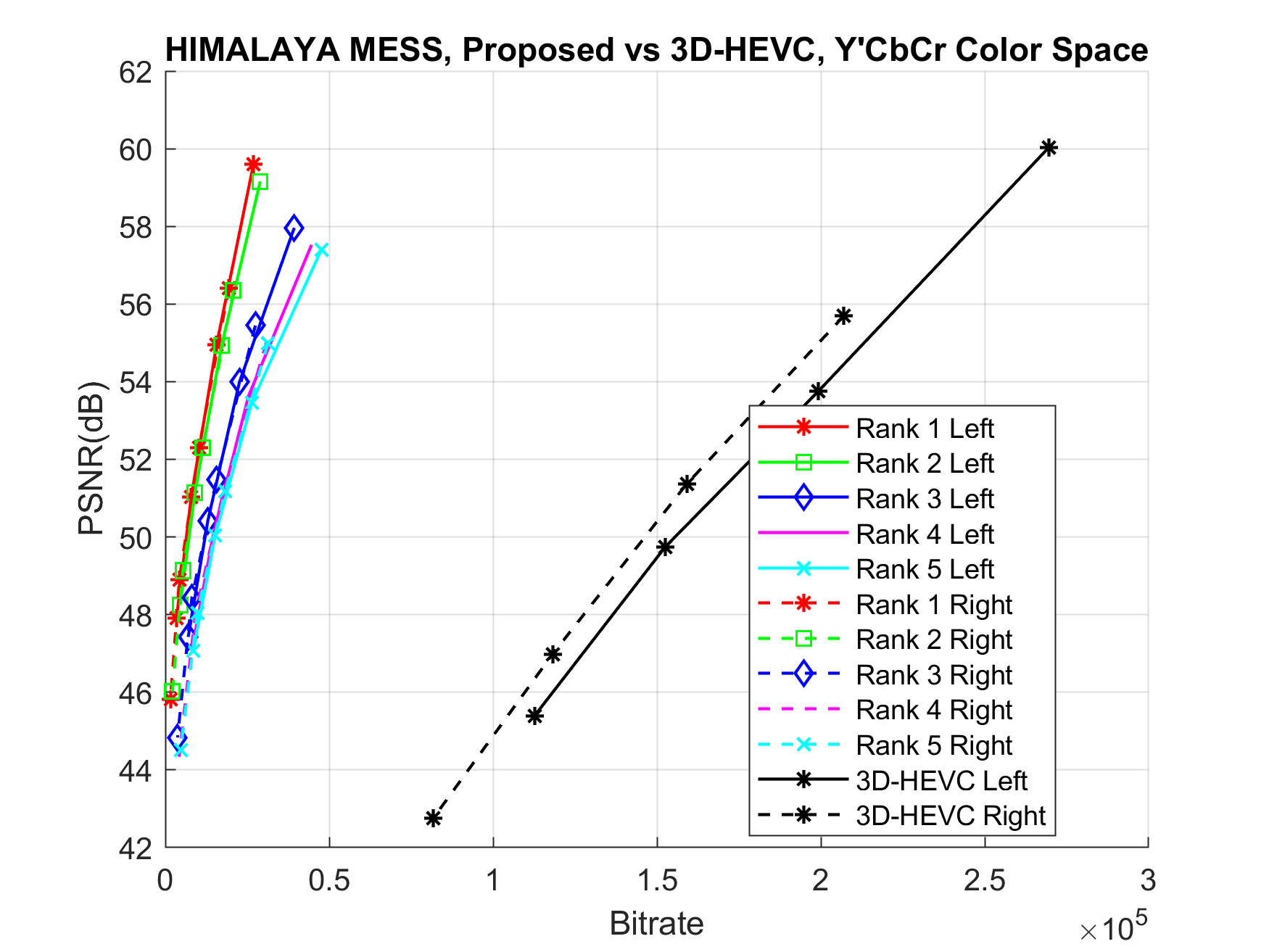}} &
\subfloat[ Room]{\includegraphics[width = 3.0in]{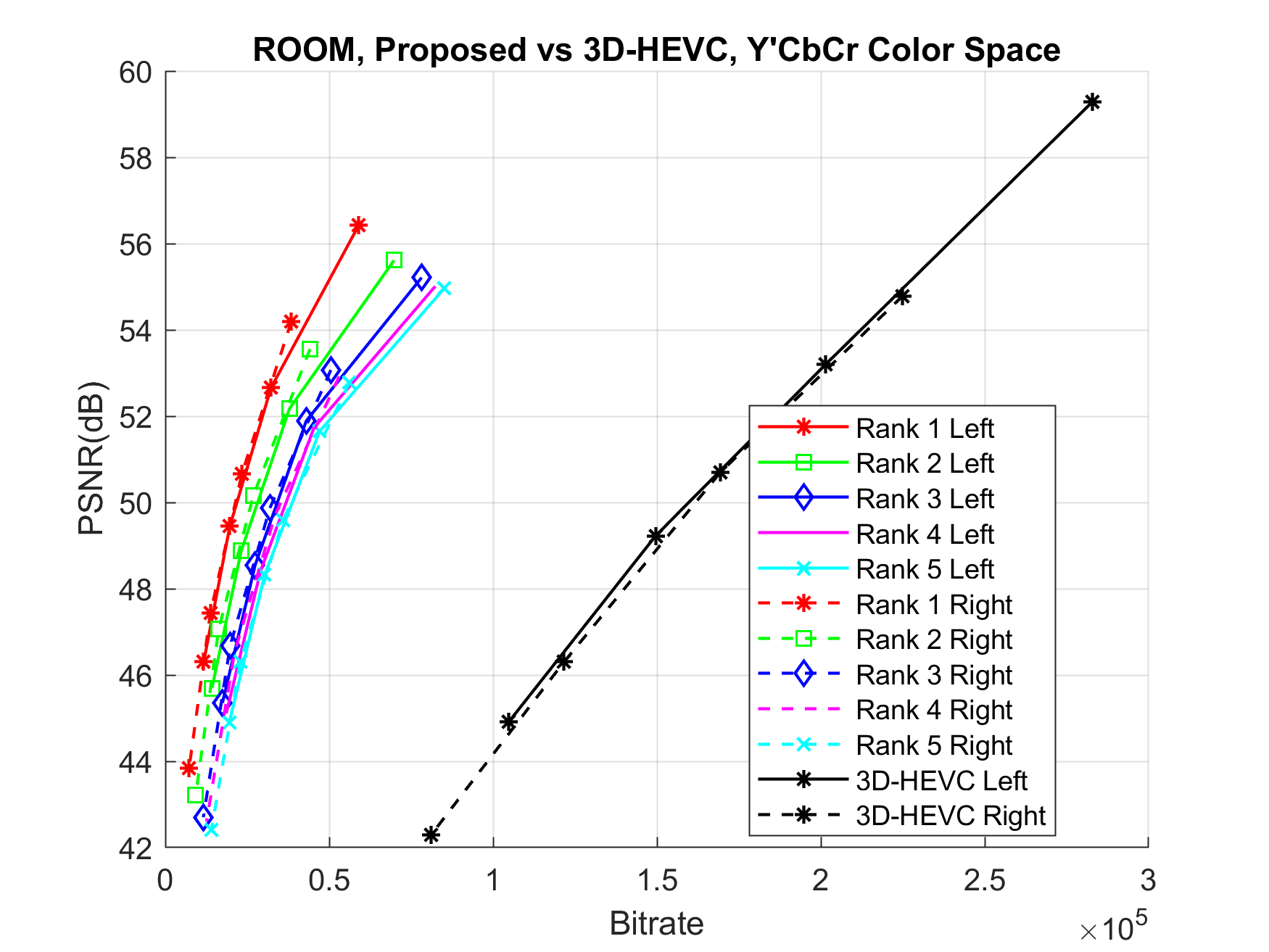}} 
\end{tabular}
\caption{Comparative analysis of proposed scheme at varying tensor rank approximation with 3D-HEVC ($Y^{'}C_bC_r$ Color Space).}
\label{Ourvs3DHEVCYCbCr}
\end{figure*}

\begin{figure*}[t] 
\centering
\begin{tabular}{cccc}
\subfloat[ Garden]{\includegraphics[width = 3.0in]{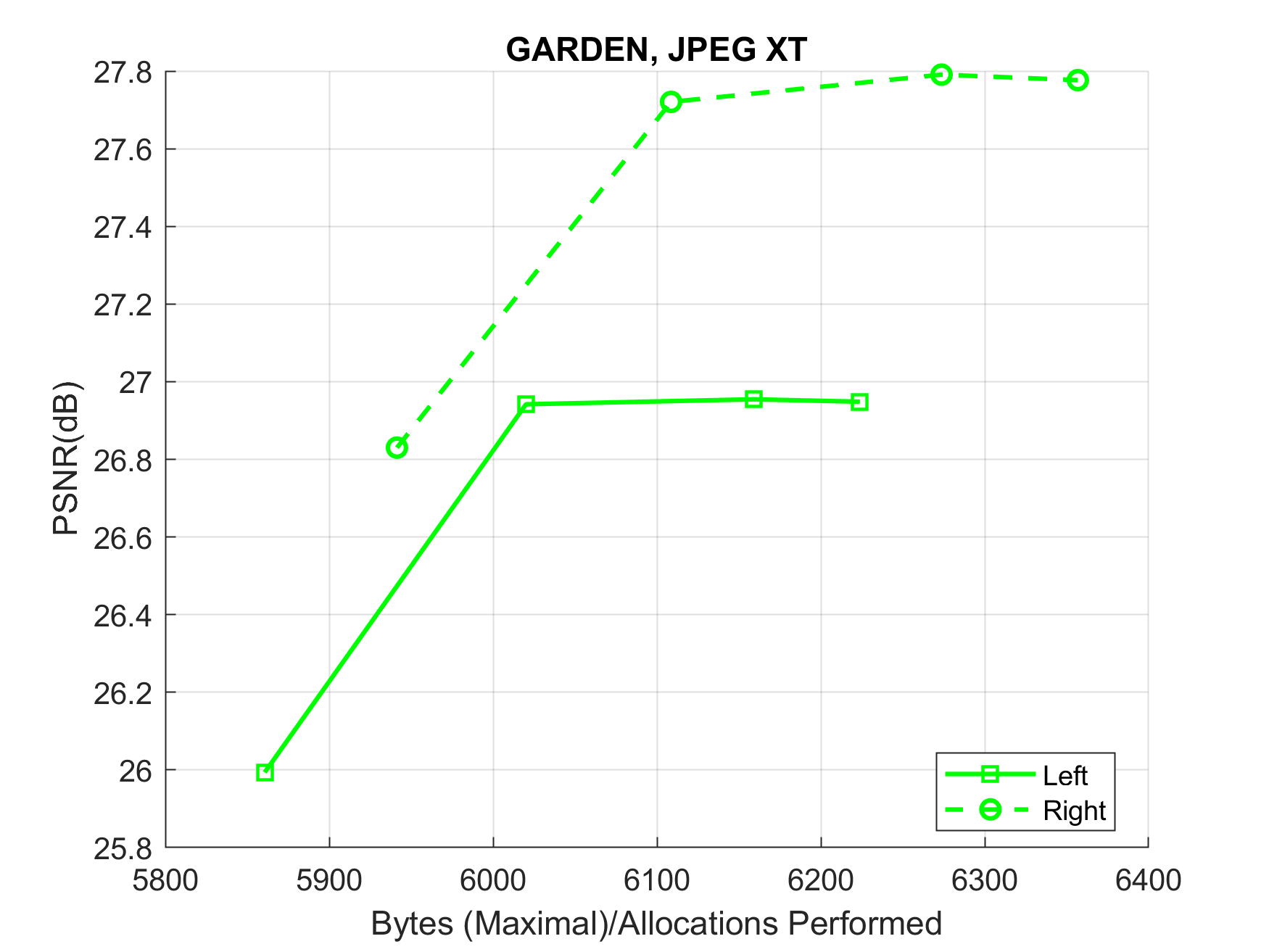}} &
\subfloat[Gate]{\includegraphics[width = 3.0in]{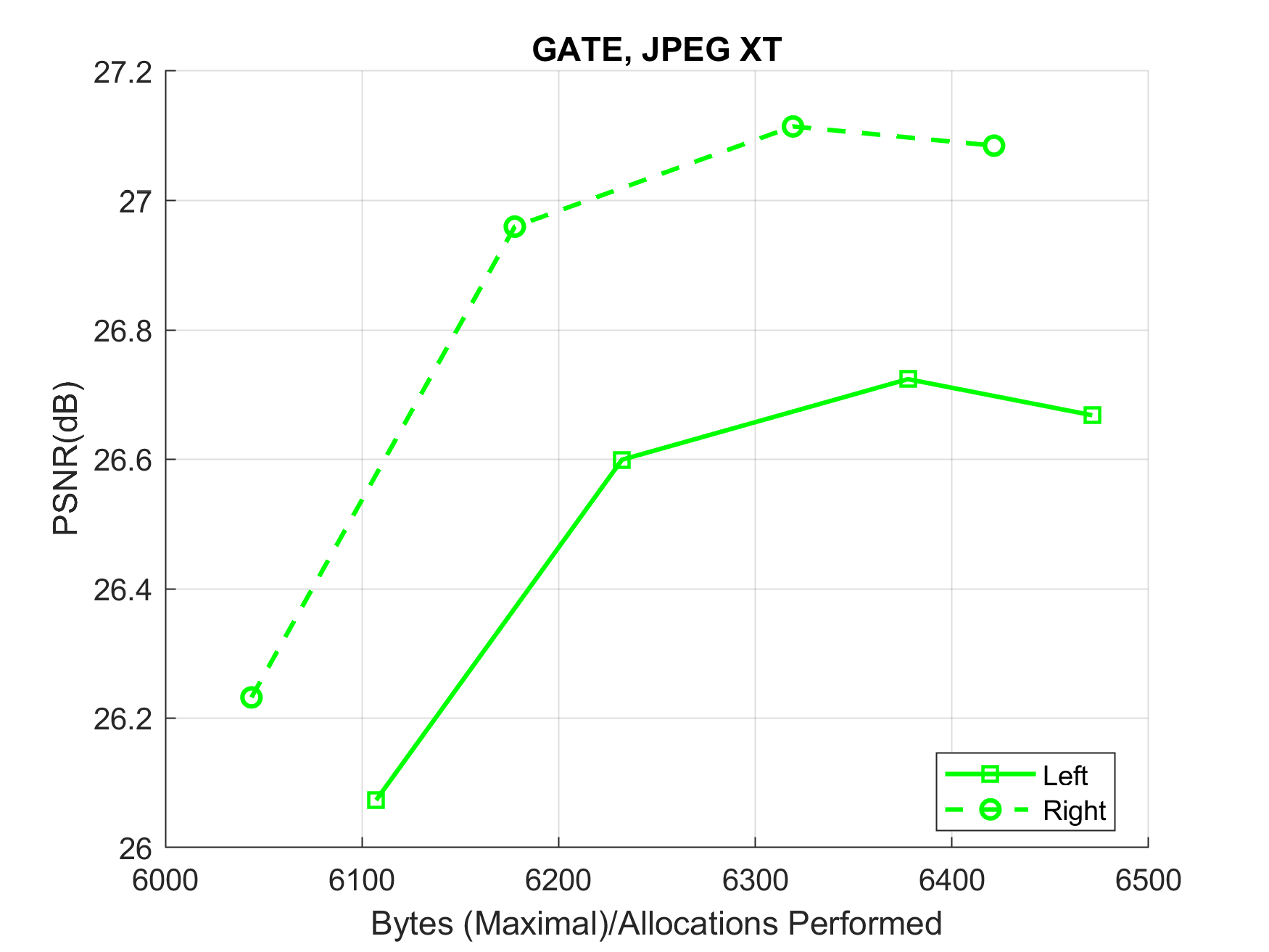}} 
 \\
\subfloat[ Himalaya Mess]{\includegraphics[width = 3.0in]{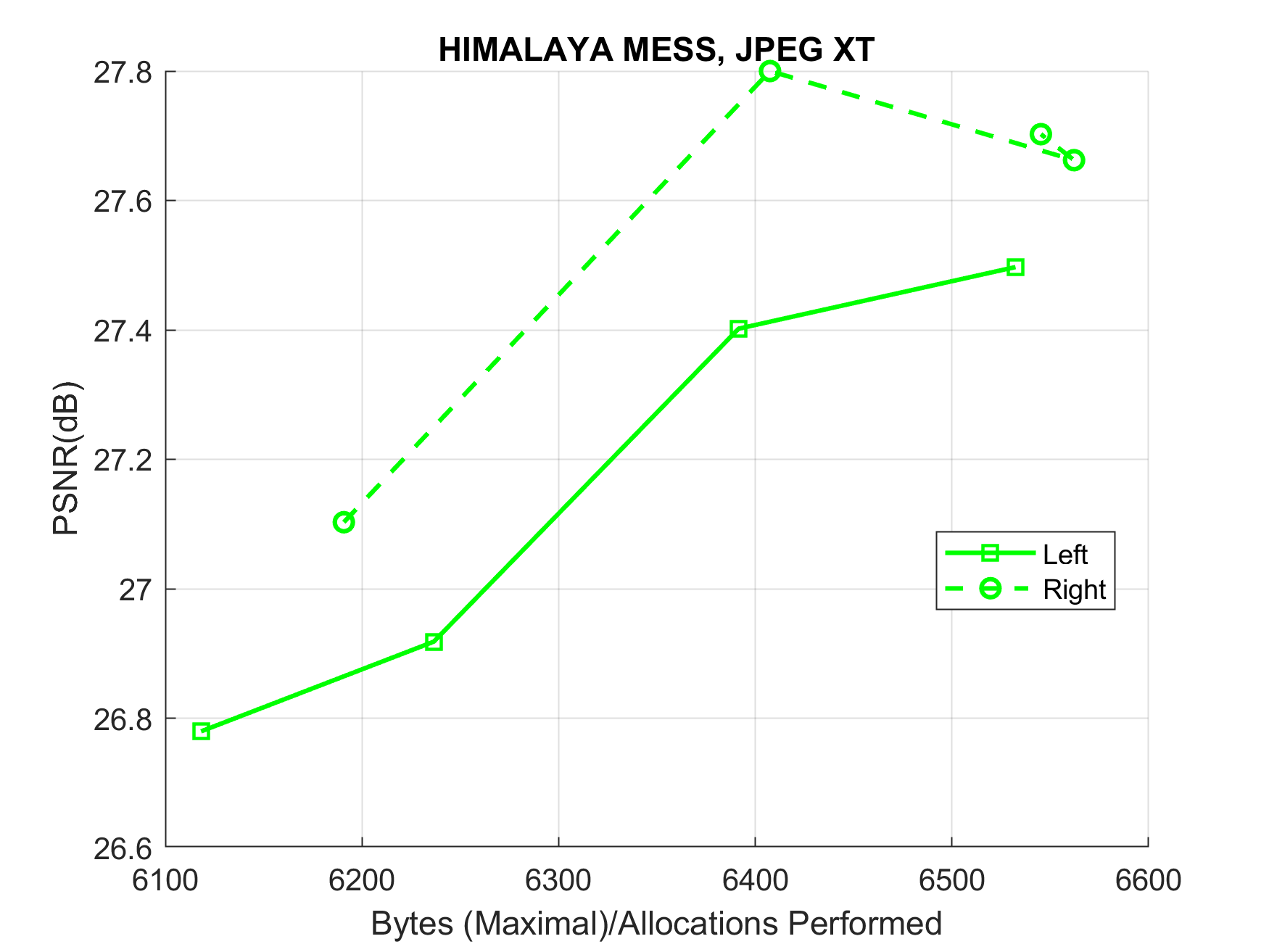}} &
\subfloat[ Room]{\includegraphics[width = 3.0in]{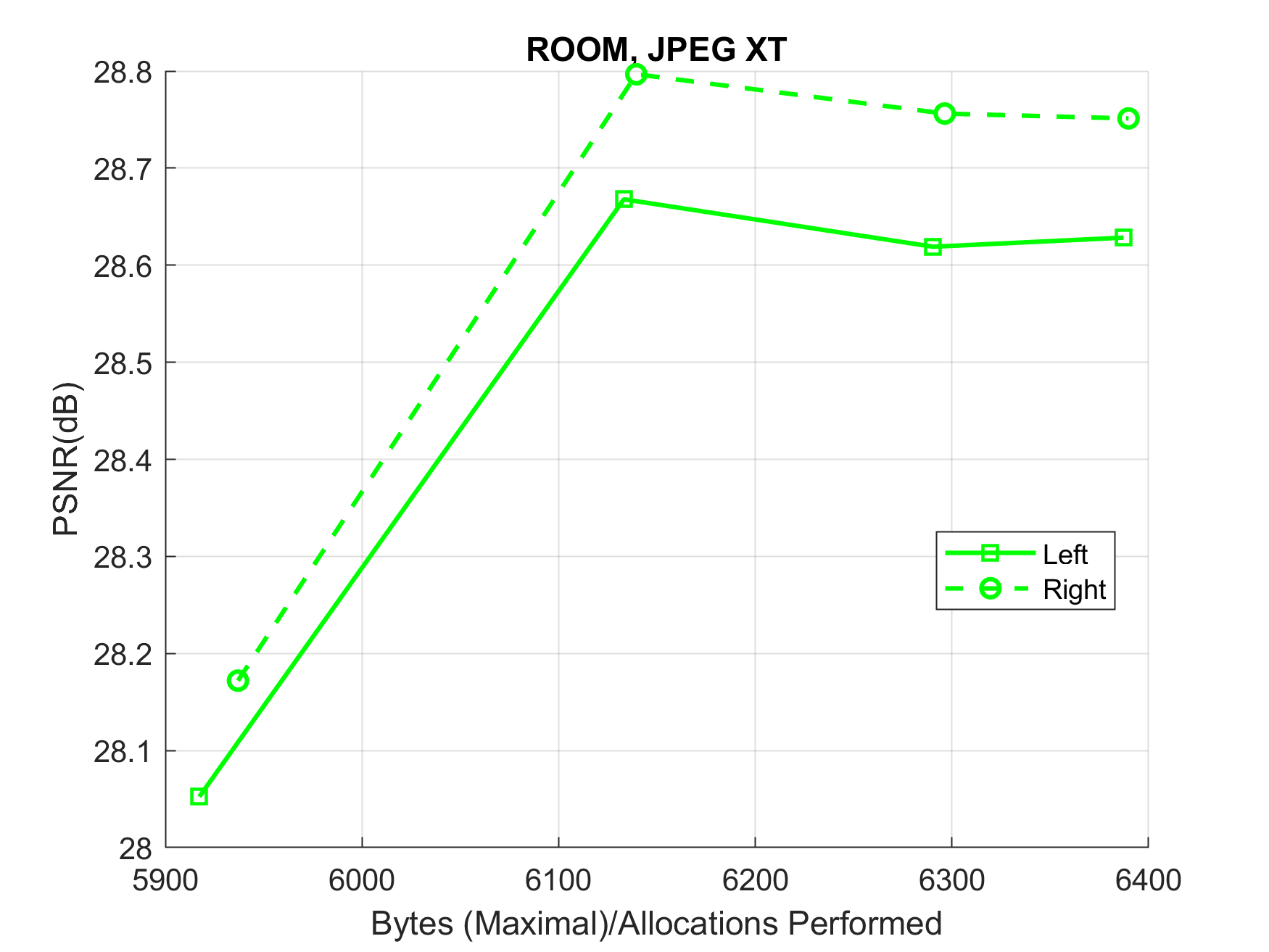}} 
\end{tabular}
\caption{Analysis of JPEG XT for coding stereo HDR images.}
\label{JPEGXTonly}
\end{figure*}
%\end{figure*}

\subsection{3D HEVC Encoding} 
The low rank approximation output $\mathfrak{\hat{T}}_{MES}^{c}$ of Tensor ALS is further encoded by 3D extension of high efficiency video coding \cite{3DHEVCRef1, 3DHEVCRef2}. Since ZED camera capture the same scene from two different viewpoints, we exploited intra-frame and inter-view and the inter-component redundancies in low-rank approximated stereo multi-exposure views to enhance the coding efficiency. 

We give low rank approximated left and right multi-exposure camera images as input to the 3D-HEVC encoder and explore the combination of inter-frame, intra-frame, and inter-view prediction. The views are not only predicted from neighboring multi-exposure images in the sequence or with data in the current frame, but also from corresponding left and right images in adjacent camera viewpoints.

The Disparity-Compensated Prediction (DCP) is basically adopted by 3D-HEVC alongside Motion-Compensated Prediction (MCP) scheme used in HEVC. In our proposed encoding framework, DCP is applied between multi-exposure left and right stereo sequences at different camera viewpoints and the MCP is applied between neighboring multi-exposure frames in the sequences respectively. The 3D-HEVC search for best block matches in the reference frames and select best candidate to generate motion disparity vector pointing to the position \cite{3DHEVCRef1, 3DHEVCRef2}. The
motion information for a current block is predicted by the motion information of one or more corresponding blocks. The corresponding blocks are located by a disparity vector in the inter-view images. This way previously encoded motion information about adjacent views is shared. The disparity vector for locating inter-view corresponding blocks could be derived by the coded disparity motion vector from a corresponding block. Lin et al. \cite{3DHEVCRef2} introduced the inter-view motion vector prediction and the inter-component motion vector prediction to improve the 3D-HEVC performance in multi-camera setting. Further, while encoding stereo multi-exposure images, the Illumination Compensation (IC) scheme is employed to adapt luminance and chrominance of inter-view predicted blocks to the current view illumination as an additional 3D-HEVC feature. The 3D-HEVC adopted linear model for parameter estimation using reconstructed samples of the current block and reference block in the prediction \cite{3DHEVCRef1}.

\begin{table*}[t]
\small
\caption{Qualitative Analysis of proposed scheme on GARDEN data (IPT color space).} 
\centering 

\begin{tabular}{|c|c|c|c|c|c|c|c|}% seven columns now, not six...
\hline
     \multicolumn{5}{|c|}{Left} & \multicolumn{3}{c|}{Right} \\
\hline
RANK & QP & BITRATE & HDR-VDP-2 & HDR-VDP-2 & BITRATE & HDR-VDP-2 & HDR-VDP-2\\
\hline
 &  &  & 24 (0.5) & 21 (0.5) & & 24 (0.5) & 21 (0.5)\\
\hline
RANK 1 & 5 & 28079.264	& 62.9677 & 61.9116 & 18711.456	& 62.1547 & 61.0311
\\
%\cline{2-8}
\hline
& 10 & 15624.384 &	60.4077	& 59.1199 & 8850.048 & 60.4074	& 59.1251
\\
\hline
%\cline{2-8}
& 15 & 7404.32	& 59.0571 &	57.6793 & 3123.648	& 59.3493 &	58.015
\\
%\cline{2-8}
\hline
& 20 &  3430.432 & 58.1302	& 56.6809 & 877.6 & 58.1049 & 56.7108
%\cline{1-8}
\\
\hline
RANK 2 & 5 &  49390.4	& 42.8382	& 40.4197 & 33251.712 &	57.4122	& 55.7724
\\
\hline
%\cline{2-8}
& 10 &  28007.008	& 62.2714	& 60.9798 & 18911.872 &	60.6781	& 59.2643 
\\
%\cline{2-8}
\hline
& 15 &  16029.088 &	59.8767	& 58.3719 & 10304.352 & 58.7761	& 57.1756
\\
%\cline{2-8}
\hline
& 20 & 9079.008 & 57.5239 & 55.8855 & 4863.616	& 56.3085 &	54.641
\\
%\cline{1-8}
\hline
RANK 3 & 5 & 63227.168	& 43.2375 &	41.3227 & 41500.736	& 51.8593 &	49.7182
\\
%\cline{2-8}
\hline
& 10 &  34594.656 &	43.8504	& 41.9826 & 24269.536	& 57.6972	& 55.6342
\\
\hline
%\cline{2-8}
& 15 & 20685.568	& 42.7912 &	40.8353 & 13985.344	& 51.6849	& 49.3394
\\
\hline
%\cline{2-8}
& 20 &  12052.896 &	42.6276	& 40.604 & 7364.064 &	54.181	& 52.1255
%\cline{1-8}
\\
\hline
RANK 4 & 5 & 68963.744	& 49.5705	& 47.8797 & 45508.608	& 42.4979	& 40.5434
\\
\hline
%\cline{2-8}
& 10 &  37336.128 &	49.9359	& 48.5535 & 26894.208 &	41.7784 & 39.6842
\\
\hline
%\cline{2-8}
& 15 &  22487.264	& 48.9767	& 47.3538 & 15820.096 &	41.6385	& 39.3831
\\
\hline
%\cline{2-8}
& 20 & 13329.056 & 40.1113	& 37.8859 & 8526.88	& 45.2914 &	42.766
%\cline{1-8}
\\
\hline
RANK 5 & 5 & 73133.376	& 50.7519 & 48.932 & 48856.096	& 43.1168 &	41.1942
\\
\hline
%\cline{2-8}
& 10 &  39480.416 &	45.7269	& 43.5288 & 29361.632	& 45.9985	& 43.606
\\
\hline
%\cline{2-8}
& 15 &  24040.928 &	49.2334	& 47.5691 & 17406.528	& 42.8599	& 40.9952
\\
\hline
%\cline{2-8}
& 20 & 14501.504 & 46.7211	& 45.1162 & 9633.472 & 45.6837 & 43.2662
%\cline{1-8}
\\
\hline
\end{tabular}
\label{GARDENOUR-QAIPT}
\end{table*}
\begin{table*}[t]
\small
\caption{Qualitative Analysis of proposed scheme on GARDEN data ($Y^{'}C_bC_r$ color space).} 
\centering 

\begin{tabular}{|c|c|c|c|c|c|c|c|}% seven columns now, not six...
\hline
     \multicolumn{5}{|c|}{Left} & \multicolumn{3}{c|}{Right} \\
\hline
RANK & QP & BITRATE & HDR-VDP-2 & HDR-VDP-2 & BITRATE & HDR-VDP-2 & HDR-VDP-2\\
\hline
 &  &  & 24 (0.5) & 21 (0.5) & & 24 (0.5) & 21 (0.5)\\
\hline

RANK 1 & 5 & 28359.328	& 65.3776	& 64.2145	&	20309.248	& 62.9982 &	61.5934
\\
%\cline{2-8}
\hline
& 10 &  16783.552	& 61.5055	& 60.1046	&	10720.864	& 59.9928	& 58.4397 
\\
%\cline{2-8}
\hline
& 15 & 	8052 &	58.7299	& 57.1152	&	3723.968	& 57.641	& 55.9066
\\
%\cline{2-8}
\hline
& 20 & 	3627.04	& 56.941 & 55.1601	&	1231.424 & 56.2498	& 54.4073
\\
%\cline{1-8}
\hline
RANK 2 & 5 & 50627.104 &	64.6524 &	63.4675	&	34754.56	& 64.1062 &	62.9188
\\
%\cline{2-8}
\hline
& 10 & 28576.864	& 61.3464	& 59.9571	&	19416.928	& 60.1247 &	58.6207
\\
%\cline{2-8}
\hline
& 15 & 	15832 &	58.7756	& 57.1551	&	10132.8	& 57.8749	& 56.1428
\\
\hline
%\cline{2-8}
& 20 & 	8580.32	& 56.4761 &	54.6907	&	4696.128	& 55.6411	& 53.7281
%\cline{1-8}
\\
\hline
RANK 3 & 5 & 64278.464 &	67.6728	& 66.7963	&	42685.376	& 65.7717 &	64.7677
\\
%\cline{2-8}
\hline
& 10 & 	35245.088	& 63.6574	& 62.4218 &		24264.96	& 61.8542	& 60.5147
\\
\hline
%\cline{2-8}
& 15 & 20481.984 &	60.024	& 58.3402	&	13471.968	& 59.0674	& 57.4403
\\
\hline
%\cline{2-8}
& 20 & 	11644.736 &	56.32	& 54.4644	&	7012.512 &	56.6161	& 54.7492
%\cline{1-8}
\\
\hline
RANK 4 & 5 & 69894.528 &	64.3 &	62.9672	&	46944.352 &	66.7841	& 65.7371
\\
\hline
%\cline{2-8}
& 10 & 	37829.312	& 62.0934	& 60.7117	&	27127.616	& 62.7161	& 61.3098
\\
\hline
%\cline{2-8}
& 15 & 	22145.44 &	59.4714	& 57.8687	&	15473.728	& 59.1619	& 57.4006
\\
\hline
%\cline{2-8}
& 20 & 12855.264 &	57.8649 &	55.9876	&	8233.44	& 56.4848	& 54.6073
%\cline{1-8}
\\
\hline
RANK 5 & 5 & 	74777.504 &	64.0669 &	62.7026	&	50129.056	& 66.6033 &	65.5509
\\
%\cline{2-8}
\hline
& 10 & 40226.624	& 64.6028	& 63.42	&	29366.528	& 62.765	& 61.4233
\\
\hline
%\cline{2-8}
& 15 & 	23715.136	& 59.5764	& 57.9333 &		16957.12	& 59.0444 &	57.3975
\\
\hline
%\cline{2-8}
& 20 & 	13942.624 &	57.5932	 & 55.6517	 &	9197.28 &	56.5437 &	54.6663
%\cline{1-8}
\\
\hline
\end{tabular}
\label{GARDENOUR-QAYCbCr}
\end{table*}

\begin{table*}[t]
\small
     \caption{Qualitative Analysis of 3D-HEVC on GARDEN data.}
     \centering
    \begin{tabular}{|c|c|c|c|c|c|c|}
    \hline
    \multicolumn{4}{|c|}{Left} & \multicolumn{3}{c|}{Right} \\
\hline
         &  BITRATE &  HDR-VDP-2 & HDR-VDP-2 &  BITRATE &  HDR-VDP-2 & HDR-VDP-2 \\
        \hline
            &   &  24 (0.5) & 21 (0.5) &  & 24 (0.5) & 21 (0.5)\\
        \hline
QP 5	& 389388.8	& 86.5589 &	86.8503 & 304436.64	& 81.4318 & 81.4073 \\
        \hline
QP 10	& 296773.92	& 79.6037 &	79.5074 & 239691.04	& 76.8393 & 76.5803 \\
        \hline
QP 15	& 236028.96	& 75.3262 & 75.0055 & 184533.12	& 72.2935 & 71.8574 \\
        \hline
QP 20	& 180811.04	& 70.6993	& 70.0197 & 133560	& 67.5243 &	66.7051 \\
        \hline
        \end{tabular}

    \label{GARDEN3DHEVC-QA}
    
\end{table*}

\begin{table*}[t]
\small
\caption{Qualitative Analysis of JPEG-XT on GARDEN data.} 
\centering

    \begin{tabular}{|c|c|c|c|c|c|c|c|c|c|}
    \hline
    \multicolumn{5}{|c|}{Left} & \multicolumn{4}{c|}{Right} \\
\hline
         &  Bytes & Allocations &  HDR-VDP-2 & HDR-VDP-2 & Bytes & Allocations &  HDR-VDP-2 & HDR-VDP-2\\
        \hline
        & Maximal & &  24 (0.5) & 21 (0.5) & Maximal  &  &  24 (0.5) & 21 (0.5)\\
        \hline
        
QP 5	& 79611008	& 13584	& 42.4481	& 40.2343 & 79376266 &	13360	& 41.6888	& 39.4845 \\

QP 10	& 79151977	& 13148	& 46.6348	& 44.4506 & 78910952	& 12918	& 46.5963 &	44.5012 \\

QP 15	& 78776787	& 12790	& 48.0697	& 45.9284 & 78483363 &	12510	& 48.7865 &	46.5013 \\

QP 20	& 78607013	& 12630	& 49.2409	& 47.0578 & 78277970	& 12314	& 49.5801 &	47.3791 \\
       \hline
        
        \end{tabular}
\label{GARDENJPEGXT-QA}
\end{table*}

\begin{table*}[t]
\small
\caption{Qualitative Analysis of proposed scheme on GATE data (IPT color space).} 
\centering 

\begin{tabular}{|c|c|c|c|c|c|c|c|}% seven columns now, not six...
\hline
     \multicolumn{5}{|c|}{Left} & \multicolumn{3}{c|}{Right} \\
\hline
RANK & QP & BITRATE & HDR-VDP-2 & HDR-VDP-2 & BITRATE & HDR-VDP-2 & HDR-VDP-2 \\
\hline
 &  &  & 24 (0.5) & 21 (0.5) & & 24 (0.5) & 21 (0.5)\\
\hline

RANK 1 & 5 & 57789.664	& 54.938 &	52.8926 & 35541.728 & 40.8112 &	38.7208
\\
%\cline{2-8}
\hline
& 10 &  30000.128	& 51.3691 &	49.1302 & 19517.184 & 47.2132	&
45.1677
\\
\hline
%\cline{2-8}
& 15 &  16961.984	& 53.0074	& 50.5973 & 10750.368	& 42.3863	& 39.9703
\\
\hline
%\cline{2-8}
& 20 &  9548.032	& 48.6316 &	46.12 & 5201.728	& 41.6219 &	39.1884
\\
%\cline{1-8}
\hline
RANK 2 & 5 &  67483.424 & 62.1948 & 60.3625 & 44507.136 & 52.3962	& 49.9829
\\
\hline
%\cline{2-8}
& 10 &  34251.392 &	51.7469	& 49.4957 & 26100.672	& 59.6466	& 57.9334
\\
\hline
%\cline{2-8}
& 15 &  20239.136	& 52.3047	& 49.8945 & 15684.704	& 55.7757	& 53.9376
\\
\hline
%\cline{2-8}
& 20 &  12085.536	& 55.7362	& 53.7246	& 8798.176 & 51.5406 & 49.0081
%\cline{1-8}
\\
\hline
RANK 3 & 5 & 74540.096	& 44.225 & 41.4202 & 47161.088	& 51.8901 & 49.462
\\
\hline
%\cline{2-8}
& 10 &  37639.776	& 61.156	& 59.6176 &	 27877.28 &	51.2004	& 48.9099
\\
\hline
%\cline{2-8}
& 15 &  22567.744	& 53.0603	& 50.4668 &	 16961.504	& 51.6594 & 49.1666
\\
\hline
%\cline{2-8}
& 20 &  13729.888	& 57.1585	& 55.4295 &	 9534.144 &	54.8893	& 52.9583
%\cline{1-8}
\\
\hline
RANK 4 & 5 & 80019.808	& 48.1616	& 45.6438	&	50446.912	& 54.9732 &	52.8886
\\
%\cline{2-8}
\hline
& 10 &  40536.992	& 53.7405	& 51.3248	&	30158.24 &	54.0148	& 51.9514
\\
%\cline{2-8}
\hline
& 15 &  24533.536	& 52.4138	& 50.2649	& 18542.752 &	58.0121	& 56.2336
\\
%\cline{2-8}
\hline
& 20 &  15059.072	& 52.0909	& 49.6655	&	10607.904	 & 55.5802	& 53.681
%\cline{1-8}
\\
\hline
RANK 5 & 5 & 84644.768	& 52.8669	& 50.7149	& 52957.344	& 52.0843	& 49.7393
\\
%\cline{2-8}
\hline
& 10 &  43008.928 &	63.5404	& 62.0053	&	31979.2	& 59.4392	& 57.5615
\\
\hline
%\cline{2-8}
& 15 &  26182.944	& 51.6787	& 49.1916	&	19722.528	& 57.6948	& 55.8014
\\
\hline
%\cline{2-8}
& 20 &  16042.912	& 51.6147	& 49.1544	&	11389.184	& 51.5317	& 49.3708
\\
%\cline{1-8}
\hline
\end{tabular}

\label{GATEOUR-QAIPT}
\end{table*}

\section{Experimental Analysis \& Results} 
The performance of the proposed scheme is tested on HDR stereoscopic image datasets \cite{Ref1, HDR3Ddataurl}. Stereo multi-exposure images/videos have been captured in \cite{Ref1} employing ZED 3D camera. The 3D HDR image dataset created by Aditya et al. \cite{Ref1} covers natural scenes. The scenes feature rich biodiversity of the Indian Institute of Technology Madras campus. The 617 acre campus adjoins Guindy National Park typify the natural range of plant and animal biodiversity in the northeastern Tamil Nadu. The stereo scene acquired under four different exposure settings. All images are acquired in 2K (full HD) resolution. The left and right views have a resolution of $2208 \times 1242$. The ZED camera is kept fixed between the successive multi-exposure captures. The images are frame-level aligned in post processing. However, captured multi-exposure images/videos of natural scenes contain slight to medium, partially traceable motion of objects. Readers refer to the publication for more details about the characteristics of the datasets \cite{Ref1}. Some stereo multi-exposure images used in the experiments are shown in Fig.~\ref{Fig2InputScenes}.

The comparative analysis is performed with JPEG-XT \cite{Ref2} and 3D-HEVC reference software 16.5 \cite{Ref3}. The JPEG-XT (ISO/IEC 18477) extends the base JPEG standard (ISO/IEC 10918-1 and ITU Rec. T.81) with support for higher integer bit depths,
high dynamic range imaging and floating-point coding. The HM-16.6 reference software of 3D-HEVC is used for evaluation \cite{Ref3}.

We generated HDR stereo images from multiple stereo photographs of the scene acquired at different exposure levels. The algorithm provided by Kalantari et al. \cite{IntroRef10} is utilized to recover HDR stereo images. In case, the motion of an object in the scene is large, we apply Wu et al. \cite{IntroRef17} learning approach to generate HDR stereo images. JPEG-XT, which supports floating point HDR image coding is then applied to the left and the right stereo images separately for comparison.  

Further, to compare coding efficiency of the proposed scheme with 3D-HEVC compression standard, we converted generated HDR stereo images into 12-bit integer version before encoding using Miller et al. \cite{Ref5} algorithm. This followed by encoding with 3D-HEVC reference software \cite{Ref3}. Table~\ref{ExperimentalPar1} summarize the experimental conditions and parameter settings with JPEG-XT and 3D-HEVC coding standards. 

We consistently achieve encouraging results with our proposed scheme of coding multi-exposure stereo images. The ``Bitrate vs PSNR'' plots shown in Fig.~\ref{Ourvs3DHEVCIPT} and Fig.~\ref{Ourvs3DHEVCYCbCr} compare proposed scheme and 3D-HEVC coding software on different scenes ``Garden'', ``Gate'', ``Himalaya Mess'', and ``Room'' respectively. We achieve significant bitrate reduction using proposed multi-exposure stereo coding scheme compared to directly encode the HDR stereo images by 3D-HEVC, while maintaining almost same quality in terms of PSNR measures. Besides, the graphs establish that the proposed formulation gives flexibility to adjust the bitrate and quality by changing the tensor rank and quantization parameters. 

Likewise, the reconstruction quality in terms of PSNR is much improved compared with JPEG-XT, while encoding at much lower bytes. The ``Bytes (maximal) vs PSNR'' plots of JPEG-XT are illustrated in Fig.~\ref{JPEGXTonly}. These plots depict the quality of decoded output vs bytes maximal required with allocations performed. Table~\ref{CompareOurIPTBytesJPEGXTAll} and Table~\ref{CompareOurYCbCrBytesJPEGXTAll} summarize maximal bytes and allocations needed to directly encode the HDR stereo images by JPEG-XT. This is compared with coding efficiency of the proposed scheme in terms of total bytes required to compress the multi-exposure stereo images under varying tensor rank settings.

The quality analysis of the proposed scheme and other state-of-the-art coders is further performed using HDR-VDP-2 algorithm \cite{Ref6}. The visibility and quality prediction is performed by generating HDR stereo images from original multi-exposure input images as well as from decoded multi-exposure output images after applying proposed coding scheme. The method \cite{IntroRef17} is used to generate HDR from LDR images. Similarly, the HDR-VDP-2 scores of 3D-HEVC and JPEG-XT reference softwares are checked, comparing input as well as decoded stereo HDR images.   

The metric prediction quality scores on all scenes are summarized in Table~\ref{GARDENOUR-QAIPT}-\ref{ROOMJPEGXT-QA}. We analysed quality scores considering diagonal display size in 24 and 21 inches, display resolution in pixels, and viewing distance of 0.5 meters. The computed quality scores of reconstructed stereo HDR images using proposed scheme are given in Table~\ref{GARDENOUR-QAIPT}, Table~\ref{GATEOUR-QAIPT}, Table~\ref{HIMALAYAOUR-QAIPT}, Table~\ref{ROOMOUR-QAIPT}, Table~\ref{GARDENOUR-QAYCbCr}, Table~\ref{GATEOUR-QAYCbCr}, Table~\ref{HIMALAYAOUR-QAYCbCr}, Table~\ref{ROOMOUR-QAYCbCr}, considering the processing of different scenes on IPT and $Y^{'}C_bC_r$ color space, respectively. In Table~\ref{GARDEN3DHEVC-QA}, Table~\ref{GATE3DHEVC-QA}, Table~\ref{HIMALAYA3DHEVC-QA}, Table~\ref{ROOM3DHEVC-QA}, quality scores of 3D-HEVC range software are provided. The HDR-VDP-2 scores of JPEG-XT on different scenes are summarized in Table~\ref{GARDENJPEGXT-QA}, Table~\ref{GATEJPEGXT-QA}, Table~\ref{HIMALAYAJPEGXT-QA}, and Table~\ref{ROOMJPEGXT-QA} respectively.

The HDR-VDP-2 metric scores clearly indicate proposed approach outperforms JPEG-XT under different test conditions. We attain the notable bitrate reduction, while maintaining appropriate reconstruction quality in terms of  HDR-VDP-2 measures, compared with the 
3D-HEVC reference software.

%\IEEEpubidadjcol
\section{Conclusion}
In this paper, a novel framework based on low rank Tensor approximation and 3D-HEVC is proposed to efficiently encode multi-exposure stereo LDR images for 3D HDR compression.  
To the best of our knowledge, this is a first encoding approach of multi-exposure stereo images based on a low-rank approximation of tensor factorization to realize HDR stereo compression. The proposed scheme offers flexibility to change the rank of core tensor and its quantization levels, and thus indisputably adapt the bitrate of stereo multi-exposure views. Moreover, covering a range of bitrates assist in taking into account distortion considering the encoding of stereo LDR multiple exposure views and in conjunction regulate the reconstruction quality of fused HDR views. This is critical to achieve stereo personalization in 3D display applications with spontaneous realism and binocular 3D depth cues \cite{3DDISPLAY1, 3DDISPLAY2, 3DDISPLAY3}. The results demonstrate that the required bitrate is much saved by the proposed scheme by efficiently exploiting the intra-frame, inter-view and the inter-component redundancies in stereo multi-exposure LDR sequences. This fulfils to achieve the goal of an integrated  model,  compatible with existing coding standards, and accommodative to maintain good quality HDR imaging in terms of PSNR and HDR-VDP-2 indices with a range of bitrates.
%%%%%%%%%%%%%%%%%%%%%%%%%%%%%%%%%%%%%%%%%%%%%%%%%%%%%%%%%%%%%%%%%%%%%%%%%%%%%%%%%%%%%

%%%%%%%%%%%%%%%%%%%%%%%%%%%%%%%%%%%%%%%%%%%%%%%%%%%%%%%%%%%%%%%%%%%%%%%%%%%%%%%%%%%%%

\begin{table*}[t]
\small
\caption{Qualitative Analysis of proposed scheme on GATE data ($Y^{'}C_bC_r$ color space).} 
\centering 

\begin{tabular}{|c|c|c|c|c|c|c|c|}% seven columns now, not six...
\hline
     \multicolumn{5}{|c|}{Left} & \multicolumn{3}{c|}{Right} \\
\hline
RANK & QP & BITRATE & HDR-VDP-2 & HDR-VDP-2 & BITRATE & HDR-VDP-2 & HDR-VDP-2\\
\hline
 &  &  & 24 (0.5) & 21 (0.5) &  & 24 (0.5) & 21 (0.5)\\
\hline

RANK 1 & 5 &  55018.624	& 51.7108 &	49.2455	&	36316.288	& 63.5339 &	62.3112
\\
%\cline{2-8}
\hline
& 10 &  29287.456	& 61.4825	& 60.1048	&	19735.808	& 54.7425 &	52.6511
\\
%\cline{2-8}
\hline
& 15 &  16002.24	& 57.5215	& 55.8076	&	9755.488 &	56.9341	& 55.2066
\\
\hline
%\cline{2-8}
& 20 & 8459.488	& 53.7522 &	51.5802		& 3740.16	& 54.8729	& 52.9268
%\cline{1-8}
\\
\hline
RANK 2 & 5 & 65617.92	& 68.4827	& 67.4344	&	44353.824 &	64.9949	& 63.8158
\\
\hline
%\cline{2-8}
& 10 & 33818.304 &	63.9766	& 62.6951	&	25387.456	& 60.8804	& 59.4889
\\
\hline
%\cline{2-8}
& 15 &  19006.144	& 60.252	& 58.6367 &		14229.216	 & 57.9942 &	56.3334
\\
\hline
%\cline{2-8}
& 20 &  10655.264	& 57.939	& 56.1044 &		7579.52	& 55.6644 &	53.7543
%\cline{1-8}
\\
\hline
RANK 3 & 5 & 73290.56	& 68.0498	 & 66.9842	&	47291.776	& 65.4025 &	64.1865
\\
\hline
%\cline{2-8}
& 10 & 37212 &	64.4951 & 	63.2797	&	27204.928	 & 55.3563	& 53.2148
\\
\hline
%\cline{2-8}
& 15 &  21445.504	& 60.5818	& 59.0821	& 15457.184 & 	58.8754	& 57.2462
\\
\hline
%\cline{2-8}
& 20 & 12224	& 57.7334 &	55.902 &	8250.848	& 56.4021	& 54.547
%\cline{1-8}
\\
\hline
RANK 4 & 5 & 79239.744 &	68.7847 &	67.7621 &	50398.56 &	65.8272 &	64.7377
\\
\hline
%\cline{2-8}
& 10 &  39920.128	& 64.3287 &	63.2048 &	29213.504	& 62.0932	& 60.7496
\\
\hline
%\cline{2-8}
& 15 &  23216.48 &	60.7778	& 59.3022	& 17041.984 &	59.1693	& 57.549
\\
\hline
%\cline{2-8}
& 20 &  13421.856	& 58.0373	& 56.2705	&	9333.536	& 56.638 &	54.7785
%\cline{1-8}
\\
\hline
RANK 5 & 5 &  84224.864 &	68.6765	& 67.6314	&	52964.352	& 65.5637 &	64.2617
\\
\hline
%\cline{2-8}
& 10 & 42476.096 &	65.216	& 64.0017	&	30887.072	& 62.2596	 & 60.854
\\
\hline
%\cline{2-8}
& 15 & 24861.92 &	61.1702	& 59.6635	&	18105.28	& 59.2304	& 57.5838
\\
\hline
%\cline{2-8}
& 20 & 14495.52 &	58.1243	& 56.3887	&	9842.208	& 56.5972 &	54.6688
\\
%\cline{1-8}
\hline

\end{tabular}

\label{GATEOUR-QAYCbCr}
\end{table*}

\begin{table*}[t]
\small
     \caption{Qualitative Analysis of 3D-HEVC on GATE data.}
     \centering

    \begin{tabular}{|c|c|c|c|c|c|c|}
    \hline
    \multicolumn{4}{|c|}{Left} & \multicolumn{3}{c|}{Right} \\
\hline
         &  BITRATE &  HDR-VDP-2 & HDR-VDP-2 &  BITRATE &  HDR-VDP-2 & HDR-VDP-2\\
        \hline
            &   &  24 (0.5) & 21 (0.5) &  & 24 (0.5) & 21 (0.5)\\
        \hline
QP 5	& 382177.76	& 88.0962	& 88.2364 & 326144	& 83.3801 &	83.3744 \\
\hline
QP 10	& 292425.76	& 81.9172	& 81.7387 & 259138.56 & 79.0632	& 78.7353\\
\hline
QP 15	& 231549.44	& 77.6057	& 77.1736 & 201603.52 &	74.4138	& 73.8644\\
\hline
QP 20	& 177331.36	& 73.2775	& 72.6117 & 148815.36	& 69.4372	& 68.7713 \\

        \hline
        \end{tabular}

    \label{GATE3DHEVC-QA}
    
\end{table*}

\begin{table*}[t]
\small
\caption{Qualitative Analysis of JPEG-XT on GATE data.} 
\centering

    \begin{tabular}{|c|c|c|c|c|c|c|c|c|c|}
    \hline
    \multicolumn{5}{|c|}{Left} & \multicolumn{4}{c|}{Right} \\
\hline
         &  Bytes & Allocations &  HDR-VDP-2 & HDR-VDP-2 & Bytes & Allocations &  HDR-VDP-2 & HDR-VDP-2\\
         \hline
         & Maximal & &  24 (0.5) & 21 (0.5) & Maximal  &  &  24 (0.5) & 21 (0.5)\\
        \hline
QP 5 &	78917261 &	12922	& 41.1913	& 38.8948 & 79089097 &	13086 &	41.2175	& 39.0037 \\

QP 10 &	78588184 &	12610	& 45.0281	& 42.695 & 78728589	& 12744	 & 45.6691	& 43.3338 \\

QP 15 &	78227658 &	12266	& 46.8118	& 44.4007 & 78370143 &	12402	& 47.6456	& 45.2561 \\

QP 20 &	78005472 &	12054	& 47.738	& 45.2956 &  78122823	& 12166 &	48.9011	& 46.5189 \\

        \hline
        
        \end{tabular}

\label{GATEJPEGXT-QA}
\end{table*}

%%%%%%%%%%%%%%%%%%%%%%%%%%%%%%%%%%%%%%%%%%%%%%%%%%%%%%%%%%%%%%%%%%%%%%%%%%%%%%%%%%%%%%%
\begin{table*}[t]
\small
\caption{Qualitative Analysis of proposed scheme on HIMALAYA MESS (IPT color space).} 
\centering 

\begin{tabular}{|c|c|c|c|c|c|c|c|}% seven columns now, not six...
\hline
     \multicolumn{5}{|c|}{Left} & \multicolumn{3}{c|}{Right} \\
\hline
RANK & QP & BITRATE & HDR-VDP-2 & HDR-VDP-2 & BITRATE & HDR-VDP-2 & HDR-VDP-2 \\
\hline
 &  &  & 24 (0.5) & 21 (0.5) &  & 24 (0.5) & 21 (0.5)\\
\hline

RANK 1 & 5 &  26054.208 & 61.2812	& 59.9823 & 18803.776	& 59.3966 & 57.8328
\\
%\cline{2-8}
\hline
& 10 &  15081.952	& 58.7169	& 57.1583	&	9442.272	& 57.4523	& 55.6999
\\
%\cline{2-8}
\hline
& 15 &  7818.752 &	56.1687	& 54.4532 & 3796.128 & 55.6676	& 53.7382
\\
%\cline{2-8}
\hline
& 20 &  3625.888 & 55.7018	& 53.9648	& 1260.032	& 55.1925	& 53.3423
%\cline{1-8}
\\
\hline
RANK 2 & 5 & 28962.336 & 68.3374	& 67.1522 & 19923.232	& 60.3525 &	59.0623
\\
%\cline{2-8}
\hline
& 10 & 17165.952 &	65.2238	& 63.8101	& 11558.688 &	57.7933 &	56.3226
\\
%\cline{2-8}
\hline
& 15 & 9026.848	& 62.806	& 61.2197	&	5896.32 &	56.0497	& 54.4638 
\\
%\cline{2-8}
\hline
& 20 & 4492.992	& 60.0746	& 58.3239	&	2753.664	& 54.297 &	52.5413
\\
%\cline{1-8}
\hline
RANK 3 & 5 &  37355.616	& 62.6863	& 61.5694	&	26972.704 &	60.1013 &	58.7841
\\
%\cline{2-8}
\hline
& 10 & 21760.896	& 59.5077 &	57.9625	&	15646.048	& 57.4343	& 55.8525
\\
%\cline{2-8}
\hline
& 15 &  12657.632	& 57.0364	& 55.3262 &		8432.192 &	55.8789	& 54.122
\\
\hline
%\cline{2-8}
& 20 & 6951.648	& 55.827 & 54.0238	&	3955.488	& 54.3115	& 52.4575
%\cline{1-8}
\\
\hline
RANK 4 & 5 & 43635.52	& 40.553 &	38.1345	&	30511.264 &	39.9455 &	37.6413
\\
%\cline{2-8}
\hline
& 10 & 24825.28	& 41.8156	& 39.4189	&	17801.6	 & 42.5748	& 40.2416
\\
%\cline{2-8}
\hline
& 15 &  14762.976 &	40.7999 &	38.4755	&	9207.904 &	42.3932	& 39.9931
\\
\hline
%\cline{2-8}
& 20 &  8159.104	& 41.6591	& 39.1767	&	4070.336 &	40.9799	 & 38.4963
%\cline{1-8}
\\
\hline
RANK 5 & 5 &  46991.68	& 42.2787 &	39.8635	&	30531.936 & 41.197 & 38.7239
\\
%\cline{2-8}
\hline
& 10 &  26204.608 & 	40.8272	& 38.2693	&	18056.352	& 53.6538 &	51.5728
\\
\hline
%\cline{2-8}
& 15 & 15637.856 & 40.8108 &	38.3558 &	9825.472 &	41.3793	& 39.0429
\\
\hline
%\cline{2-8}
& 20 &  8979.456 & 52.2592	 & 50.0298	&	4768.416 &	53.2353 &	51.3023
\\
%\cline{1-8}
\hline
\end{tabular}

\label{HIMALAYAOUR-QAIPT}
\end{table*}

\begin{table*}[t]
\small
\caption{ Qualitative Analysis of proposed scheme on HIMALAYA MESS ($Y^{'}C_bC_r$ color space).
} 
\centering 

\begin{tabular}{|c|c|c|c|c|c|c|c|}% seven columns now, not six...
\hline
     \multicolumn{5}{|c|}{Left} & \multicolumn{3}{c|}{Right} \\
\hline
RANK & QP & BITRATE & HDR-VDP-2 & HDR-VDP-2 & BITRATE & HDR-VDP-2 & HDR-VDP-2\\
\hline
 &  &  & 24 (0.5) & 21 (0.5) & & 24 (0.5) & 21 (0.5)\\
\hline

RANK 1 & 5 & 26884.256	& 65.1586	& 64.1123 &	19330.656	& 62.1582	& 60.9067
\\
\hline
%\cline{2-8}
& 10 & 15656.16	& 61.7995 &	60.5225	& 10253.28	& 58.964 &	57.4596
\\
\hline
%\cline{2-8}
& 15 & 	7893.824	& 59.1368	& 57.6652	&	4290.4 &	56.8391 &	55.2158
\\
\hline
%\cline{2-8}
& 20 & 	3420.832	& 56.7765	& 55.1215	&		1575.328 &	55.0928	& 53.3057
%\cline{1-8}
\\
\hline
RANK 2 & 5 & 28962.336 &	68.3374	& 67.1522	&	20787.136 &	66.17	& 64.8307
\\
\hline
%\cline{2-8}
& 10 & 17165.952 &	65.2238	& 63.8101	&	11380.896 &	63.5429	 & 61.9929
\\
\hline
%\cline{2-8}
& 15 & 9026.848 &	62.806	& 61.2197	&	5372.928	& 61.3182 &	59.6177
\\
%\cline{2-8}
\hline
& 20 & 	4492.992	& 60.0746 &	58.3239	&	2164.384 &	59.0335	& 57.2385
\\
%\cline{1-8}
\hline
RANK 3 & 5 & 39339.36 &	49.372	& 46.7206	&	27458.944 &	67.1737 &	65.8461
\\
\hline
%\cline{2-8}
& 10 & 22619.36	& 57.9869	& 55.7046	&	15645.984 &	64.2727 &	62.7338
\\
\hline
%\cline{2-8}
& 15 & 	12955.648	& 50.2996	& 47.9479	&	8117.312	& 61.6355	& 59.8942
\\
\hline
%\cline{2-8}
& 20 & 	6987.84	& 60.4275 &	58.6148	&	3721.056	& 58.8431	& 56.8775
%\cline{1-8}
\\
\hline
RANK 4 & 5 & 44609.952	& 53.9785	& 52.0577	&	30855.328	& 67.1596 &	65.9635
\\
\hline
%\cline{2-8}
& 10 & 25382.08	& 58.7247 &	56.9575	&	17742.08 &	64.1094	& 62.6511
\\
\hline
%\cline{2-8}
& 15 & 14660.608	& 54.2983	& 52.2783 &		9193.888 &	61.5485	& 59.8551
\\
\hline
%\cline{2-8}
& 20 & 8047.68	& 60.8144 &	59.1127	&	4203.328	& 58.9818 &	57.0852
%\cline{1-8}
\\
\hline
RANK 5 & 5 & 47599.328 &	61.5714 &	59.7028	&	31300.096	& 67.5924 &	66.4386
\\
\hline
%\cline{2-8}
& 10 & 26339.264	& 59.0427	& 56.8585 &		18242.496	& 64.4047	& 62.8976
\\
\hline
%\cline{2-8}
& 15 & 15241.376	& 60.9151 &	59.1247 &		9735.84	 & 62.0198	& 60.3219
\\
\hline
%\cline{2-8}
& 20 & 8505.76	& 60.9219	& 59.1987	&	4680.8 &	59.258 &	57.3323
%\cline{1-8}
\\
\hline

\end{tabular}

\label{HIMALAYAOUR-QAYCbCr}
\end{table*}

\begin{table*}[t]
\small
     \caption{Qualitative Analysis of 3D-HEVC on HIMALAYA MESS data.}
     \centering
    \begin{tabular}{|c|c|c|c|c|c|c|}
    \hline
    \multicolumn{4}{|c|}{Left} & \multicolumn{3}{c|}{Right} \\
\hline
         &  BITRATE &  HDR-VDP-2 & HDR-VDP-2 &  BITRATE &  HDR-VDP-2 & HDR-VDP-2\\
        \hline
          &   &  24 (0.5) & 21 (0.5) &  & 24 (0.5) & 21 (0.5)\\
        \hline
QP 5	& 269528	& 85.3871	& 85.5883 & 206973.92 &	81.6204	& 81.7454 \\
\hline
QP 10	& 199202.4	& 79.4406	& 79.4143 & 159073.28 &	76.3242	& 76.159\\
\hline
QP 15	& 152535.04	& 74.9644	& 74.6324 & 118164.8 & 71.5489	& 71.0437\\
\hline
QP 20	& 112638.4	& 70.3373	& 69.7498 & 81748 &	67.047	& 66.2815\\

        \hline
        \end{tabular}

    \label{HIMALAYA3DHEVC-QA}
    
\end{table*}

\begin{table*}[t]
\small
\caption{Qualitative Analysis of JPEG-XT on HIMALAYA MESS data.} 
\centering

    \begin{tabular}{|c|c|c|c|c|c|c|c|c|c|}
    \hline
    \multicolumn{5}{|c|}{Left} & \multicolumn{4}{c|}{Right} \\
\hline
         &  Bytes & Allocations &  HDR-VDP-2 & HDR-VDP-2 & Bytes & Allocations &  HDR-VDP-2 & HDR-VDP-2\\
        \hline
        & Maximal & &  24 (0.5) & 21 (0.5) & Maximal  &  &  24 (0.5) & 21 (0.5)\\
        \hline
QP 5	& 78887787	& 12894	& 40.73	& 38.6356 & 78697127 & 12712 & 40.9958 &	38.8447 \\

QP 10	& 78579670	& 12600	& 45.4976	& 43.344 &  78158474	& 12198	& 46.2007 & 	44.0688 \\

QP 15	& 78193993	& 12234	& 47.0491	& 44.8717 & 77800058 &	11856	& 47.7803 & 	45.6275 \\

QP 20	& 77864923	& 11920	& 48.9214	& 46.8068 & 77835690 &	11892	& 49.3209 &	47.0741\\
        \hline
        
        \end{tabular}

\label{HIMALAYAJPEGXT-QA}
\end{table*}

%%%%%%%%%%%%%%%%%%%%%%%%%%%%%%%%%%%%%%%%%%%%%%%%%%%%%%%%%%%%%%%%%%%%%%%%%%%%%%%%%%%%%%%%

\begin{table*}[t]
\small
\caption{Qualitative Analysis of proposed scheme on ROOM data (IPT color space).} 
\centering 

\begin{tabular}{|c|c|c|c|c|c|c|c|}% seven columns now, not six...
\hline
     \multicolumn{5}{|c|}{Left} & \multicolumn{3}{c|}{Right} \\
\hline
RANK & QP & BITRATE & HDR-VDP-2 & HDR-VDP-2 & BITRATE & HDR-VDP-2 & HDR-VDP-2 \\
\hline
 &  &  & 24 (0.5) & 21 (0.5) &  & 24 (0.5) & 21 (0.5)\\
\hline

RANK 1 & 5 &  58999.52	& 48.1087 &	45.8614 & 38131.968 & 41.4502 &	39.3074
\\
%\cline{2-8}
\hline
& 10 &  32356.896	& 57.0954	& 54.7766	& 24033.088 &	41.1792	& 38.9312
\\
%\cline{2-8}
\hline
& 15 &  20364.832 &	48.0147	& 45.2258	& 14441.216	& 38.8711	& 36.6038
\\
%\cline{2-8}
\hline
& 20 &  12617.504 &	43.9524	& 41.2823	&	7909.152	& 39.8297	& 37.6275
\\
%\cline{1-8}
\hline
RANK 2 & 5 &  71821.152	& 60.4676	& 58.8533	&	45482.816 &	44.2458 &	41.6944
\\
%\cline{2-8}
\hline
& 10 & 39619.488 &	51.7287	& 49.6298	&	28561.888	& 52.6594 &	50.195 
\\
%\cline{2-8}
\hline
& 15 &  25076.096 &	46.3318	& 44.0273	&	17777.984	& 52.9636	& 50.6605
\\
%\cline{2-8}
\hline
& 20 &  15831.456 &	45.3535 &	42.618 &	10536.96 & 	44.2459 &	41.5996
%\cline{1-8}
\\
\hline
RANK 3 & 5 &  77940.096	& 42.9619	& 40.3653	&	50035.232	& 52.5723 &	50.4242
\\
%\cline{2-8}
\hline
& 10 &  42962.336 &	43.4708	& 40.9031	& 31728.032	& 59.3287	& 57.6333
\\
\hline
%\cline{2-8}
& 15 & 27254.944 &	45.9587	& 43.7813	&	19966.272 &	54.194	 & 52.2353
\\
%\cline{2-8}
\hline
& 20 & 17181.536 &	42.8521	& 40.2609	&	11932.032 &	53.6955	 & 51.4613
%\cline{1-8}
\\
\hline
RANK 4 & 5 &  81955.104	& 59.6374	& 58.0407	&	52553.056	& 59.5554 &	58.0785
\\
%\cline{2-8}
\hline
& 10 & 45381.664 &	55.8806	& 53.8359	&	33578.304	& 54.8063	& 52.8356 
 \\
%\cline{2-8}
\hline
& 15 &  29373.888 &	49.6271	& 47.3292	&	21299.328 &	44.2287 &	41.5356
\\
%\cline{2-8}
\hline
& 20 &  18743.424	& 49.5191	& 47.2103	&	12844.16	& 53.6704	& 51.6061
%\cline{1-8}
\\
\hline
RANK 5 & 5 &  84626.176	& 51.4125 &	49.0315	&	55664.448 &	55.4891 &	53.4093
\\
\hline
%\cline{2-8}
& 10 &  47189.856	& 57.488 &	55.2905	&	35923.36 &	51.0514	 & 48.7444
\\
\hline
%\cline{2-8}
& 15 &  30665.344 &	53.564	& 51.16	&	23142.72 &	52.7341	& 50.461
\\
\hline
%\cline{2-8}
& 20 &  19651.936	& 47.8211	& 45.3837	&	14305.056	& 53.1723	& 51.1185
\\
%\cline{1-8}
\hline
\end{tabular}

\label{ROOMOUR-QAIPT}
\end{table*}

\begin{table*}[t]
\small
\caption{Qualitative Analysis of proposed scheme on ROOM data ($Y^{'}C_bC_r$ color space).
} 
\centering 
\begin{tabular}{|c|c|c|c|c|c|c|c|}% seven columns now, not six...
\hline
     \multicolumn{5}{|c|}{Left} & \multicolumn{3}{c|}{Right} \\
\hline
RANK & QP & BITRATE & HDR-VDP-2 & HDR-VDP-2 & BITRATE & HDR-VDP-2 & HDR-VDP-2\\
\hline
 &  &  & 24 (0.5) & 21 (0.5) & & 24 (0.5) & 21 (0.5)\\
\hline

RANK 1 & 5 & 59025.536	& 67.1934	& 66.1516 &	38415.648	& 63.7828	& 62.604
\\
%\cline{2-8}
\hline
& 10 & 	32253.056	& 62.5784 &	61.2813	&	23366.816	& 59.7357	& 58.1907
\\
%\cline{2-8}
\hline
& 15 & 19675.904	& 58.9333 &	57.3447 &	13751.68 &	56.1695 &	54.2317
\\
\hline
%\cline{2-8}
& 20 & 	11721.024 &	55.9963	& 54.138	&	7253.728	& 53.9449	& 51.9819
%\cline{1-8}
\\
\hline
RANK 2 & 5 & 	69779.808	& 68.4256 &	67.5247	&	44165.472 &	64.7369	& 63.6076
\\
\hline
%\cline{2-8}
& 10 & 37910.656	 &	66.1075 & 65.1253	&	26857.824	& 62.214 &	60.8567
\\
\hline
%\cline{2-8}
& 15 & 23180.032 &	61.0723	& 59.6918	& 16275.648 &	58.1212 &	56.4347
\\
%\cline{2-8} 
\hline
& 20 & 14228.064 &	58.5716	& 56.9465	&	9283.584 &	54.9396 &	53.0424
%\cline{1-8}
\\
\hline
RANK 3 & 5 & 78188.16	& 69.7134	& 68.9295	&	50512.384	& 66.0846 &	65.004
\\
%\cline{2-8}
\hline
& 10 & 43021.504	& 65.4703 &	64.4271	&	31878.272	& 62.3091	& 60.9601
\\
\hline
%\cline{2-8}
& 15 & 	27422.048 &	61.8632	& 60.5324	&	19796.192	& 59.192 &	57.529
\\
\hline
%\cline{2-8}
& 20 & 	17423.552	& 58.742	& 57.1198	&	11605.952	& 55.6046 &	53.6836
%\cline{1-8}
\\
\hline
RANK 4 & 5 & 82350.976 &	70.5289	& 69.7775	&	53027.136	& 66.0044
& 64.9649
\\
\hline
%\cline{2-8}
& 10 & 45312.8	& 65.9683	& 64.8792	&	33514.752	& 62.1869	& 60.8606
\\
\hline
%\cline{2-8}
& 15 & 29074.688	& 62.6891	& 61.3974	&	21044.192	& 58.8571	& 57.1992
\\
\hline
%\cline{2-8}
& 20 & 18595.776	& 59.3915 &	57.768	&	12577.184 &	55.5701	 & 53.6435
%\cline{1-8}
\\
\hline
RANK 5 & 5 & 85088.128	& 70.3894 &	69.6352	&	56181.056	& 66.4325	& 65.4513
\\
\hline
%\cline{2-8}
& 10 & 	47055.52 &	66.1075	& 65.1253	& 35923.392 &	62.214	& 60.8567
\\
\hline
%\cline{2-8}
& 15 & 	30302.752 &	62.6898	& 61.4704	&	22868.576	& 59.0155 &	57.3384
\\
\hline
%\cline{2-8}
& 20 & 	19530.272 &	59.3602	& 57.741	&	14061.856	& 55.787	& 53.7779

%\cline{1-8}
\\
\hline

\end{tabular}

\label{ROOMOUR-QAYCbCr}
\end{table*}

\begin{table*}[t]
\small
     \caption{Qualitative Analysis of 3D-HEVC on ROOM data.}
     \centering
    \begin{tabular}{|c|c|c|c|c|c|c|}
    \hline
    \multicolumn{4}{|c|}{Left} & \multicolumn{3}{c|}{Right} \\
\hline
         &  BITRATE &  HDR-VDP-2 & HDR-VDP-2 &  BITRATE &  HDR-VDP-2 & HDR-VDP-2\\
        \hline
        &   &  24 (0.5) & 21 (0.5) &  & 24 (0.5) & 21 (0.5)\\
        \hline    
QP 5 &	282889.92	& 84.677 &	84.8065 & 224956.32	& 80.4038	& 80.2779\\
\hline
QP 10 &	201493.44	& 78.8156 &	78.6143 & 169375.68	& 75.9265	& 75.5515 \\
\hline
QP 15 &	149564.96	& 74.3842 &	73.9334 & 121671.84	& 70.9176	& 70.3211 \\
\hline
QP 20 &	104762.88	& 69.3825 &	68.671 & 81078.24 & 65.8374 &	64.9497 \\

        \hline
        \end{tabular}

    \label{ROOM3DHEVC-QA}
    
\end{table*}

\begin{table*}[t]
\small
\caption{Qualitative Analysis of JPEG-XT on ROOM data.} 
\centering

    \begin{tabular}{|c|c|c|c|c|c|c|c|c|c|}
    \hline
    \multicolumn{5}{|c|}{Left} & \multicolumn{4}{c|}{Right} \\
\hline
         &  Bytes & Allocations &  HDR-VDP-2 & HDR-VDP-2 & Bytes  & Allocations &  HDR-VDP-2 & HDR-VDP-2\\
        \hline
        & Maximal & &  24 (0.5) & 21 (0.5) & Maximal  &  &  24 (0.5) & 21 (0.5)\\
        \hline
QP 5 &	79445323 &	13426 &	43.0581	& 40.8897 & 79388716 &	13372 &	42.5443 &	40.462 \\
\hline
QP 10 &	78847959 &	12856 &	47.6063	& 45.5043 & 78831182	& 12840 &	48.0341 &	45.911 \\
\hline
QP 15 &	78441324 &	12470 &	49.6562	 & 47.5443 & 78426642	& 12456 &	50.1281 &	48.0604 \\
\hline
QP 20 &	78204475 &	12244 &	50.326	 & 48.2997 & 78198185	& 12238	& 51.0012 &	48.929 \\
            \hline
        
        \end{tabular}
  
\label{ROOMJPEGXT-QA}
\end{table*}

%%%%%%%%%%%%%%%%%%%%%%%%%%%%%%%%%%%%%%%%%%%%%%%%%%%%%%%%%%%%%%%%%%%%%%%%%%%%%%%%%%%

%%%%%%%%%%%%%%%%%%%%%%%%%%%%%%%%%%%%%%%%%%%%%%%%%%%%%%%%%%%%%%%%%%%%%%%%%%%%%%%%%

\begin{table*}[t]
\small
     \caption{Comparison of JPEG-XT and Proposed scheme (IPT color space).}
     \centering
    
\begin{tabular}{|c|c|c|p{1.5cm}|p{1.5cm}|c|c|c|c|c|}% seven columns now, not six...
\hline
    SCENE & \textbf{QP}  & \multicolumn{2}{|c|}{JPEG XT} & \multicolumn{5}{c|}{PROPOSED} \\
%\cline{3-10}
\hline
& & Bytes (Maximal) & Allocations & RANK 1 &  RANK 2 &  RANK 3 &  RANK 4 &  RANK 5  \\
\hline
GARDEN & 5 & 	158987274	& 26944	&	1462268	& 2582624	& 3272805 &	3577319 &	3812229
\\
\hline
%\cline{2-10}
& 10  & 158062929	& 26066 &	764884	& 1466273 &	1839564	 & 2007256	& 2151372
\\
\hline
%\cline{2-10}
& 15 &	157260150	& 25300	& 	329057	& 822978	& 1083524	& 1197163	& 1295291
\\
\hline
%\cline{2-10}
& 20 &	156884983	& 24944	&	134684	& 435765	& 606838	& 683056	& 754276
\\
\hline
%\cline{2-10}
%\hline
GATE & 5 &	158006358	& 26008		& 2916664 &	3499763	& 3803220 &	4077143	& 4300124
\\
\hline
%\cline{2-10}
& 10 &  157316773	& 25354		& 1547474	& 1886060	& 2047466 &	2209284 &	2343437
\\
\hline
%\cline{2-10}
& 15 & 	156597801	& 24668	&	866069	& 1122678	& 1235347 &	1346192 &	1434604
\\
\hline
%\cline{2-10}
& 20 &  156128295	& 24220	&	460988	& 652674	& 727059	& 802151	& 857311
\\
%\cline{2-10}
\hline
%\hline
HIMALAYA & 5 & 157584914	& 25606		& 1401870 &	1527732	& 2010318 &	2317145 &	2422671
\\
%\cline{2-10}
\hline
& 10 &  156738144 &	24798	&	766440	& 897703 &	1169025 &	1332148 &	1383213
\\
%\cline{2-10}
\hline
& 15 & 	155994051	& 24090		& 363023	& 466407 &	659115 &	749148 &	795787
\\
%\cline{2-10}
\hline
& 20 &  155700613 &	23812	&	152743	& 226516	& 340906 &	382228	& 429679
\\
%\cline{2-10}
\hline
ROOM & 5 &  158834039	& 26798	&	3035417	& 3665807	& 3999287	& 4203438	& 4384140
\\
%\cline{2-10}
\hline
& 10 & 157679141	& 25696		& 1762245	& 2130726	& 2334132	& 2467557 &	2597346
\\
\hline
%\cline{2-10}
& 15 &  156867966	& 24926	&	1087747	& 1339248 &	1475721	& 1583596	& 1681560
\\
\hline
%\cline{2-10}
& 20 &  156402660	& 24482	&	641516	& 824071	& 909857 &	987170	& 1061214
\\
%\cline{2-10}
\hline
\end{tabular}
    
    \label{CompareOurIPTBytesJPEGXTAll}
    
\end{table*}

\begin{table*}[t]
\small
     \caption{Comparison of JPEG-XT and Proposed scheme ($Y^{'}C_bC_r$ color space).}
     \centering
    \begin{tabular}{|c|c|c|p{1.5cm}|p{1.5cm}|c|c|c|c|c|}% seven columns now, not six...
\hline
    SCENE & \textbf{QP} & \multicolumn{2}{|c|}{JPEG XT} & \multicolumn{5}{c|}{PROPOSED} \\
%\cline{3-10}
\hline
&  & Bytes (Maximal) & Allocations & RANK 1 &  RANK 2 &  RANK 3 &  RANK 4 &  RANK 5  \\
\hline
GARDEN & 5 &  	158987274	& 26944	&	1520951	& 2668235	& 3342678	& 3651273	& 3903388
\\
%\cline{2-10}
\hline
& 10 &  158062929	& 26066		& 859571 &	1499864	& 1859747 &	2029962	& 2174844
\\
%\cline{2-10}
\hline
& 15 & 	157260150	& 25300	&	368057	& 811458 &	1061119	& 1175657	& 1271066
\\
%\cline{2-10}
\hline
& 20 &  156884983 &	24944	&	151885 &	414947 &	583097 &	659080 &	723180
\\
%\cline{2-10}
\hline
GATE & 5  & 158006358 &	26008	&	2854274	& 3436675	& 3768256	& 4051255	& 4287221
\\
\hline
%\cline{2-10}
& 10 &  	157316773	& 25354	&	1532035	& 1850238	& 2013087 &	2160484 &	2292657
\\
\hline
%\cline{2-10}
& 15 & 	156597801	& 24668	&	804987 &	1038663	& 1153267 &	1258135 &	1342783
\\
\hline
%\cline{2-10}
& 20 & 	156128295	& 24220	&	381297	& 569895	& 639897	& 711164 &	760612
\\
%\cline{2-10}
\hline
HIMALAYA & 5 & 157584914	& 25606	&	1444274 &	1554729 &	2087505	& 2358348	& 2465665
\\
\hline
%\cline{2-10}
& 10 &  156738144	& 24798	&	809728	& 892147	& 1195850	& 1347688 &	1393238
\\
\hline
%\cline{2-10}
& 15 &  155994051	& 24090	&	380815 &	450051	& 658588	& 745511	& 780596
\\
\hline
%\cline{2-10}
& 20 & 	155700613 &	23812		& 156188	& 208101	& 334711	& 382902 &	412138
%\cline{2-10}
\\
\hline
ROOM & 5 & 158834039	& 26798	&	3045095	& 3560848	& 4021950 &	4230624	& 4414720
\\
%\cline{2-10}
\hline
& 10 & 	157679141	& 25696		& 1738179	& 2024073	& 2340676	& 2463419 &	2593149
\\
%\cline{2-10}
\hline
& 15 &	156867966	& 24926	&	1044670	& 1233048	& 1475628	& 1566273	& 1661662
\\
\hline
%\cline{2-10}
& 20 &  156402660	& 24482	&	593019	& 734797 &	907230	& 974213	& 1049812
%\cline{2-10}
\\
\hline
\end{tabular}

    \label{CompareOurYCbCrBytesJPEGXTAll}
    
\end{table*}

%%%%%%%%%%%%%%%%%%%%%%%%%%%%%%%%%%%%%%%%%%%%%%%%%%%%%%%%%%%%%%%%%%%%%%%%%%%%%%%%%%%%%
\bibliographystyle{ieeetr}
\footnotesize
\bibliography{root.bib}

\end{document}